\pdfoutput=1
\documentclass[11pt]{article}

\usepackage[final]{acl}
\usepackage{afterpage}
\usepackage{cuted}
\usepackage{times}
\usepackage{latexsym}
\usepackage{listings}
\usepackage{graphicx}
\usepackage{subcaption}
\usepackage[T1]{fontenc}
\usepackage[utf8]{inputenc}
\usepackage{microtype}
\usepackage{inconsolata}
\usepackage{graphicx}
\usepackage{booktabs}  

\usepackage{latexsym}
\usepackage{adjustbox}
\usepackage{xspace}
\usepackage{todonotes}
\usepackage{xcolor}
\usepackage{comment}
\usepackage{tikz}
\usepackage{pgfplots}
\usepackage{pgfplotstable}
\usepackage{makecell}
\usepackage{adjustbox}
\usepackage{pifont}
\usepackage{float}
\usepackage{supertabular}
\usepackage{multirow}
\usepackage{longtable}
\usepackage{graphicx}
\usepackage[edges]{forest}
\definecolor{hidden-draw}{RGB}{0,0,0}
\usepackage{wrapfig}
\usepackage{CJKutf8}
\usepackage{CJK}

\usepackage{tcolorbox}
\tcbset{
    mybox/.style={
        fontupper=\linespread{.8}\selectfont,
        sharp corners,
        top=0pt, 
        bottom=0pt 
    }
}

\usepackage{amsmath}
\usepackage{amssymb}

\usepackage{booktabs}
\usepackage{threeparttable}
\usepackage{tabularx}

\usepackage{soul}

\usepackage{bm} 

\usepackage{algorithm}
\usepackage{algorithmic}

\usepackage{setspace}

\definecolor{mygreen}{RGB}{11,141,10}
\definecolor{myred}{RGB}{223,68,52}
\definecolor{myblue}{RGB}{70,130,180}
\definecolor{mydeepblue}{RGB}{65,105,225}
\definecolor{myviolet}{RGB}{97,0,138}
\definecolor{myburgundy}{RGB}{110,10,30}
\definecolor{myblue2}{RGB}{0,105,148}
\definecolor{iceblue}{RGB}{173, 216, 230}
\definecolor{puregreen}{RGB}{0, 218, 0}
\definecolor{graygreen}{RGB}{74,113,106}

\definecolor{wingreen}{rgb}{0,0.45,0.24}
\definecolor{losered}{rgb}{1.0,0.1,0.24}

\definecolor{lightcoral}{rgb}{0.97, 0.36, 0.46}
\definecolor{lightyellow}{rgb}{0.98, 0.7, 0}
\definecolor{harvestgold}{rgb}{0.85, 0.57, 0.0}
\definecolor{brightlavender}{rgb}{0.75, 0.58, 0.89}
\definecolor{capri}{rgb}{0.0, 0.75, 1.0}
\definecolor{carminepink}{rgb}{0.92, 0.3, 0.26}
\definecolor{celadon}{rgb}{0.67, 0.88, 0.69}
\definecolor{darkpastelgreen}{rgb}{0.01, 0.75, 0.24}

\definecolor{grayhighlight}{RGB}{250,250,227}

\definecolor{target}{HTML}{F47983}
\definecolor{control}{HTML}{3E87CD}
\definecolor{credibility}{HTML}{B98AC9}
\definecolor{logical}{HTML}{93C572}
\definecolor{emotional}{HTML}{F9EAC3}

\newenvironment{packeditemize}{
\begin{list}{$\bullet$}{
\setlength{\labelwidth}{8pt}
\setlength{\itemsep}{0pt}
\setlength{\leftmargin}{\labelwidth}
\addtolength{\leftmargin}{\labelsep}
\setlength{\parindent}{0pt}
\setlength{\listparindent}{\parindent}
\setlength{\parsep}{0pt}
\setlength{\topsep}{3pt}}}{\end{list}}

\usepackage{pifont}
\newcommand{\crossmark}{\text{\ding{55}}}
\renewcommand{\checkmark}{\text{\ding{51}}}

\newcommand{\diffdown}[1]{\raisebox{0.5pt}{\fontsize{6}{5.5}\selectfont{\textcolor{wingreen}{\textbf{$\blacktriangledown${$ #1$}}}}}}
\newcommand{\diffup}[1]{\raisebox{0.5pt}{\fontsize{6}{5.5}\selectfont{\textcolor{losered}{\textbf{$\blacktriangle${$ #1$}}}}}}

\newcommand{\myline}{\par
  \kern0pt 
  \hrule height 0.6pt
  \kern3pt 
}

\newcommand{\mylinenoskip}{\par
  \kern3pt 
  \hrule height 0.6pt
  \kern3pt 
}

\usepackage{amsthm}

\usepackage{url}
\usepackage{booktabs}
\usepackage{siunitx}
\title{Stop Before You Fail: Operational Capability Boundaries for Mitigating Unproductive Reasoning in Large Reasoning Models}

\author{
    \centerline{Qingjie Zhang\textsuperscript{1}, \ 
    Yujia Fu\textsuperscript{1}, \ 
    Yang Wang\textsuperscript{2}, \ 
    Liu Yan\textsuperscript{2},
    } \vspace{0.5mm}  \\
    \centerline{\textbf{
    Tao Wei\textsuperscript{2}, \
    Ke Xu\textsuperscript{1}, \ 
    Minlie Huang\textsuperscript{1}, \ 
    and Han Qiu\textsuperscript{1*}
    }} \vspace{0.5mm} \\
    \centerline{\normalsize{$^{1}$Tsinghua University, $^{2}$Ant Group}} \vspace{0.5mm} \\
    \centerline{\normalsize{Emails: \{qj-zhang24@mails., qiuhan@\}tsinghua.edu.cn~~\textsuperscript{*}Corresponding author}}
}

\begin{document}
\maketitle
\begin{abstract} 
Current answering paradigms for Large Reasoning Models (LRMs) often fail to account for the fact that some questions may lie beyond the model’s operational capability boundary, leading to long but unproductive reasoning. In this paper, we study whether LRMs expose early signals predictive of such cases, and whether these signals can be used to mitigate unproductive reasoning. In black-box settings, we find that reasoning expressions contain failure-predictive signals. In white-box settings, we show that the hidden states of the last input token contain information that is predictive of whether a question will not be solved correctly under our evaluation setup. Building on these observations, we propose two test-time monitoring strategies: reasoning expression monitoring and hidden states monitoring, that reduce token usage by 62.7–93.6\%, substantially improving efficiency and reliability while largely preserving accuracy. We open-source \href{https://github.com/qingjiesjtu/CapBound}{our code}.
\end{abstract}

\section{Introduction}

Large Reasoning Models (LRMs) have demonstrated remarkable capabilities on complex reasoning tasks such as mathematics \citep{guo2025deepseek,jaech2024openai,ahn2024large}. However, when confronted with hard questions, they frequently exhibit unproductive reasoning, manifested as repetitive looping or error accumulation until context limits are exhausted \citep{zhou2025gsm,chen2025towards,yao2025understanding,mukherjee2025premise} (as shown in \autoref{app:output}).

This phenomenon suggests an important limitation of current answering paradigms: they often fail to account for the fact that some questions are likely to fall beyond the model's \emph{operational capability boundary} under a fixed evaluation setup, leading to long yet ultimately unproductive reasoning traces. If such cases can be identified early, reasoning could become more reliable and efficient: instead of continuing low-value deliberation, the model may abstain from full reasoning and provide a concise partial approach.

\begin{figure*}
    \centering
    \includegraphics[width=0.99\linewidth]{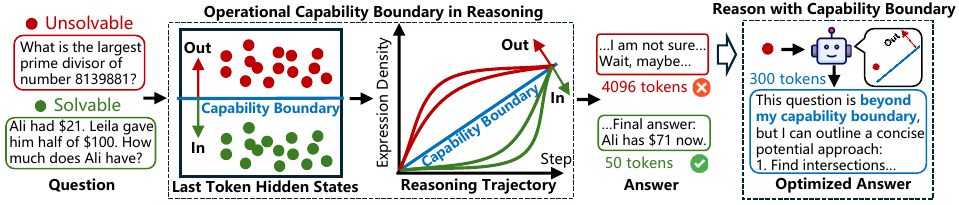}
    \caption{Signals related to the operational capability boundary are reflected in reasoning trajectories as expression density patterns, and in the last token hidden states even before reasoning begins. Leveraging these signals enables LRMs to identify questions that are likely to remain unsolved and to provide more reliable and efficient responses.}
        \label{fig:overview}
    \vspace{-1em}
\end{figure*}

Existing approaches to studying reasoning limits have primarily taken an external perspective, for example by constructing increasingly challenging benchmarks such as Humanity's Last Exam (HLE) \citep{phan2025humanity}, or by designing real-time routers that estimate question difficulty and route harder questions to more capable models \citep{openai2025gpt5}. In contrast, relatively little work has examined whether such limits are reflected in the model's internal signals, despite the importance of this question for both understanding reasoning behavior and further improving reliability and efficiency at test time.

Accordingly, this paper studies whether LRMs expose \textbf{\textit{early signals predictive of eventual failure}}, and whether such signals can be used to mitigate unproductive reasoning. We use the term \emph{capability boundary} in an operational sense: under a fixed evaluation setup, it refers to the empirical separation between questions the model eventually solves and those it does not. \autoref{fig:overview} provides an overview of our approach. From both black-box and white-box perspectives, we identify signals that help distinguish unsolvable from solvable questions. Leveraging these signals, we enable LRMs to assess, on the fly, whether a question is likely to remain unsolved under the current setup, thereby improving the efficiency and reliability of reasoning. We summarize our contributions as:

\begin{packeditemize}
\item We provide empirical evidence that LRMs' reasoning expressions are associated with the correctness of final answers, revealing language patterns that are predictive of eventual failure.
\item From a black-box perspective, we characterize failure-predictive reasoning trajectories. We identify a diverging pattern: an accelerated growing confidence trajectory for solved questions, but a convergent uncertainty trajectory for unsolved questions.
\item From a white-box perspective, we show that the hidden states of the last input token contain information predictive of whether a question will eventually be solved correctly. In particular, solvable and unsolvable questions are close to linearly separable in the hidden space, even before reasoning begins.
\item We design two test-time monitoring strategies: reasoning expression monitoring (black-box) and hidden states monitoring (white-box). Experiments show that these strategies substantially improve efficiency and reliability while largely preserving accuracy, reducing token usage by $62.7$--$93.6\%$.
\end{packeditemize}

\section{Related Work}

\noindent \textbf{From knowledge boundary to capability boundary.} Recent studies have examined the knowledge boundaries of Large Language Models (LLMs), focusing on the limits of factual knowledge encoded in parameters \citep{yin2024benchmarking,li2024knowledge,wen2024perception,deng2025unveiling}. These works primarily emphasize factual coverage, with retrieval-augmented generation (RAG) being a common solution to extend the accessible knowledge space \citep{gao2023retrieval,lewis2020retrieval}.

By contrast, the capability boundary concerns what reasoning tasks LLMs can reliably solve \citep{chen2025rbf++,xue2025reliablemath,chen2024unlocking}. This notion is broader yet rarely formalized. Existing efforts typically approximate it externally: either by constructing more challenging benchmarks such as Humanity’s Last Exam (HLE) \citep{phan2025humanity}, or by designing routing mechanisms such as GPT-5 router that delegate harder questions to more capable models \citep{openai2025gpt5}. \textit{In contrast, we examine whether capability boundaries are reflected in models’ internal signals.}

\vspace{0.5em}
\noindent \textbf{Interpretability of reasoning expressions and hidden states.} Interpretability aims to uncover how LLMs understand and execute tasks, providing insights that enhance their reliability and efficiency. Recent studies have approached from both reasoning expressions and hidden states. On the reasoning expressions side, \citet{guo2025deepseek} highlight the anthropomorphic cues, ``aha moments'', as a key role in reasoning. \citet{qian2025demystifying} tracks mutual information along the generation and finds sharp surges at specific thinking tokens (e.g., ``Hmm'', ``Wait'', ``Therefore''). \citet{bogdan2025thought} frame sentence-level importance as a counterfactual influence problem, showing that only a few critical utterances steer the reasoning trajectory.

On the hidden states side, \citet{wendler2024llamas} probe multilingual Llama models and reveal English as the pivot language in the internal structure. \citet{fan2024not} demonstrate that certain intermediate layers already encode sufficient predictive features, implying redundancy in the full forward pass. \citet{zhang2025reasoning} train linear probes on hidden states to verify intermediate answers, anticipating future correctness. \textit{Building on both perspectives, our work analyzes reasoning expressions and hidden states to study signals related to the operational capability boundary.}

\vspace{0.5em}
\noindent \textbf{Efficient reasoning: from solvable questions to unsolvable questions.} A range of recent works examine the phenomenon of overthinking \citep{sui2025stop,hou2025thinkprune}. \citet{chen2024not} show that LRMs often allocate excessive computation to easy questions, while \citet{zhang2024understanding} report that multi-turn self-correction can lead to unnecessary cost.
To mitigate such inefficiency, \citet{han2025token} compresses unnecessary reasoning by including a reasonable token budget in the prompt. \citet{wang2025wait} demonstrate that suppressing frequent reflective tokens such as ``Wait'', ``Hmm'', or ``Alternatively'' improves reasoning efficiency without harming accuracy. \citet{yang2025dynamic,fu2025deep,fu2024efficiently} introduce mechanisms to decide during reasoning whether further steps are unnecessary, allowing the model to terminate early and output the final answer directly.

However, these works focus on cases where LRMs can solve a question but still waste computation by reasoning for too long. By contrast, \textit{our work addresses a complementary and less-studied scenario: questions that are likely to fall beyond the model’s operational capability boundary, where LRMs may nevertheless continue reasoning unproductively despite being unlikely to arrive at a correct answer under the current setup.}
\section{LRMs Expose Signals of Failure}
\label{sec:cannotSolve}

We begin our exploration by evaluating GPT-oss-20B \citep{agarwal2025gpt}, DeepSeek-R1-0528-Qwen3-8B (abbreviated as R1-Distill-Qwen3-8B) \citep{guo2025deepseek}, DeepSeek-R1-Distill-Qwen-32B (abbreviated as R1-Distill-Qwen-32B) \citep{guo2025deepseek}, and QwQ-32B \citep{qwq-32b} on several standard mathematical benchmarks, including AIME’25 \citep{aime2025}, AIME’24 \citep{aime2024}, HMMT \citep{hmmt_feb_2025}, and AMC’23 \citep{amc2023_10a} (see \autoref{app:modelsAndDataset} for detailed descriptions of datasets and models). 

\begin{figure}
    \includegraphics[width=\linewidth]{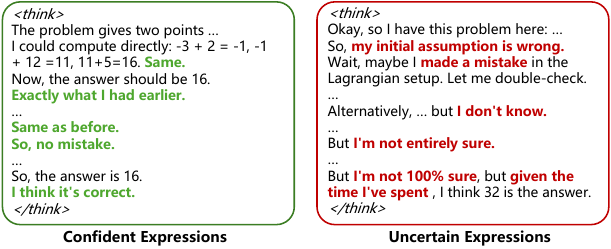}
    \vspace{-2ex}
    \caption{Confident vs. uncertain expressions. 
    }
    \label{fig:expressionReasoning}
    \vspace{-3ex}
\end{figure}

Inspired by works which monitors LRMs' reasoning \citep{baker2025monitoring,he2025can,luo2024improve}, we examine the reasoning processes (e.g., the segments wrapped within \verb|<think>| in DeepSeek-like models) to look for internalized assessments regarding whether the input question can be correctly solved. As illustrated in \autoref{fig:expressionReasoning}, we observe that such signals, expressed as anthropomorphic tones \citep{guo2025deepseek,yang2025understanding}, emerge throughout the reasoning process, and become especially evident in its final stages. Toward the end of reasoning, models often produce explicit statements that reflect their confidence level. Strikingly, these expressed beliefs tend to align with the correctness of their final answers. For example, when the reasoning contains more confident expressions such as ``So, no mistake'' or ``I think it's correct'', the final answers are typically correct. Conversely, when uncertain expressions like ``My initial assumption is wrong'' or ``I'm not 100\% sure” appear, the final answers are incorrect.


This observation motivates us to quantify the alignment between expressed belief and answer correctness, which we formalize through:
\begin{equation}
\begin{aligned}
\text{Can \%} &= \mathbb{P}(\checkmark \mid \text{More confidence}), \\
\text{Cannot \%} &= \mathbb{P}(\crossmark \mid \text{More uncertainty}).
\end{aligned}
\end{equation}
Specifically, \text{Can \%} is the proportion of questions for which the model expresses more confidence during reasoning and indeed produces a correct answer, while \text{Cannot \%} is the proportion of questions for which the model expresses more uncertainty and indeed fails to provide the correct answer.

\begin{table}
\centering
\resizebox{0.85\linewidth}{!}{
\begin{tabular}{l|c|c}
\toprule
                & \textbf{Can \%} & \textbf{Cannot \%} \\
\midrule
GPT-oss-20B            & $80.2 $  & $100$  \\
R1-Distill-Qwen3-8B    & $97.6$  & $100$  \\
R1-Distill-Qwen-32B    & $90.6$  & $100$  \\
QwQ-32B                & $86.0$  & $100$  \\
\bottomrule
\end{tabular}
}
\vspace{-1ex}
\caption{LRMs’ final answers aligns with their expressed belief, especially for incorrect final answers.}
\label{tab:indication}
\vspace{-3ex}
\end{table}

\autoref{tab:indication} shows the results. We observe that LRMs' expressed beliefs align with the correctness of their final answers. In particular, for questions where LRMs express more uncertainty during reasoning, the final answers are consistently incorrect (\text{Cannot \%} = 100). This phenomenon is observed across multiple models, suggesting that \textbf{\textit{LRMs expose signals of failure}}.

\begin{figure*}
    \centering
    \includegraphics[width=0.95\linewidth]{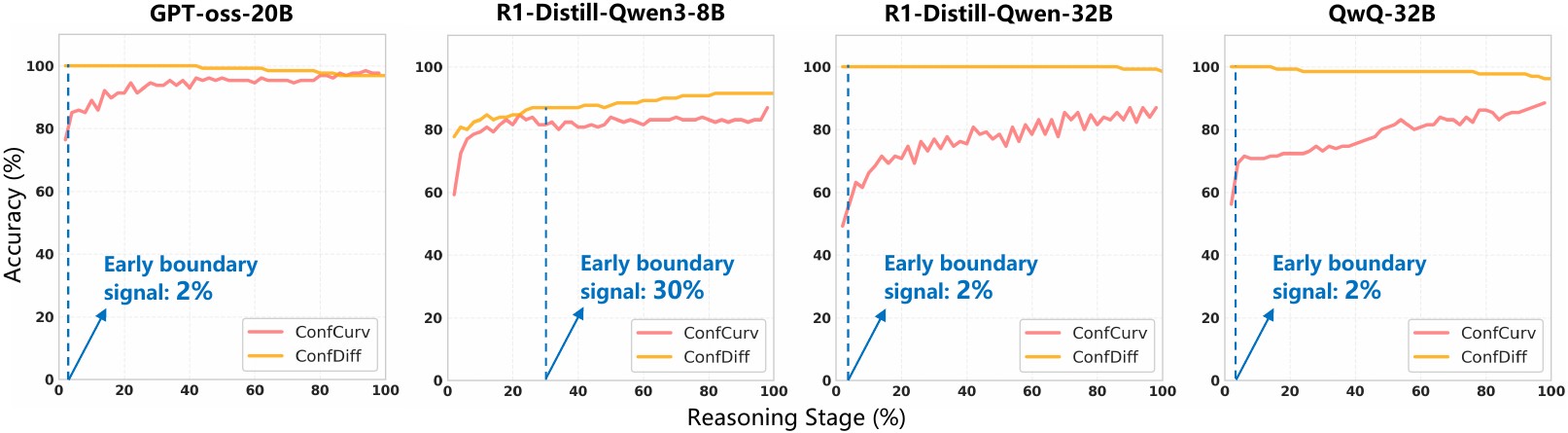}
    \vspace{-0.5em}
    \caption{Accuracy (\%) of ConfDiff and ConfCurv to separate unsolvable and solvable questions through out the reasoning stage. ConfDiff is more reliable as it achieves higher accuracy and exhibits smaller fluctuations. Boundary signal appears at an early stage (e.g., 2\%).
    }
    \label{fig:confDiffCurv}
    \vspace{-1.5em}
\end{figure*}

\section{Capability Boundary in Expressions}
\label{sec:cbReasoningTokens}

Motivated by the observations in the previous section, we hypothesize that LRMs' reasoning expressions may contain signals related to the operational capability boundary. We therefore systematically investigate the dynamics of confident and uncertain expressions throughout the reasoning process.

To ensure that the analysis is not tied to any particular LRM's reasoning style, we first extract confident and uncertain expressions separately for each model, based on its characteristic reasoning patterns (see the detailed extraction pipeline and phrase-list sensitivity analysis in \autoref{app:confidentUncertainExpression}). Such anthropomorphic expressions are commonly observed in LRM reasoning traces and have been discussed in prior work as informative features of the reasoning process \citep{yang2025understanding,guo2025deepseek,qian2025demystifying}.

Then, we uniformly divide the reasoning process into multiple stages (e.g., 50 stages) and compute the density of confident and uncertain expressions at each stage. Finally, concatenating these values yields the expression density trajectories, denoted as
$\mathcal{D}_C(t)$ for confident expressions and $\mathcal{D}_U(t)$ for uncertain expressions. These trajectories capture how the density of each type of expression evolves over the course of reasoning.

\begin{figure}[t]
    \centering
    \includegraphics[width=0.9\linewidth]{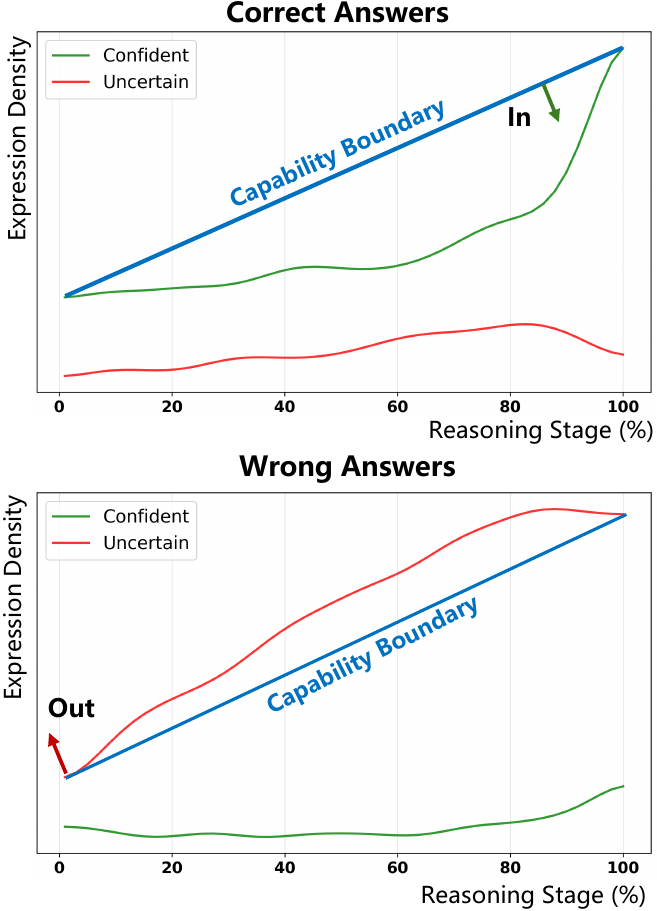}
    \vspace{-0.5em}
    \caption{Eventually solved questions show concave confident-expression curves, whereas unsolved questions show convex uncertain-expression curves.}   
        \label{fig:CBtoken}
    \vspace{-1.5em}
\end{figure}

We plot $\mathcal{D}_C(t)$ and $\mathcal{D}_U(t)$ in \autoref{fig:CBtoken}, separately for cases with correct and wrong final answers. We observe that LRMs exhibit clear differences in their reasoning expression trajectories depending on final answer correctness. When the final answer is correct, the confident trajectory lies above the uncertain trajectory and shows an accelerating increasing trend. In contrast, when the final answer is incorrect, the uncertain trajectory dominates over the confident trajectory and tends to converge. This contrast suggests that reasoning expression trajectories contain useful signals (blue line in \autoref{fig:CBtoken}) for distinguishing questions that are eventually solved from those that remain unsolved under the current setup. In particular, the dominant trajectory tends to be more concave for solvable questions and more convex for unsolvable ones.

\begin{figure*}
    \centering
    \includegraphics[width=0.9\linewidth]{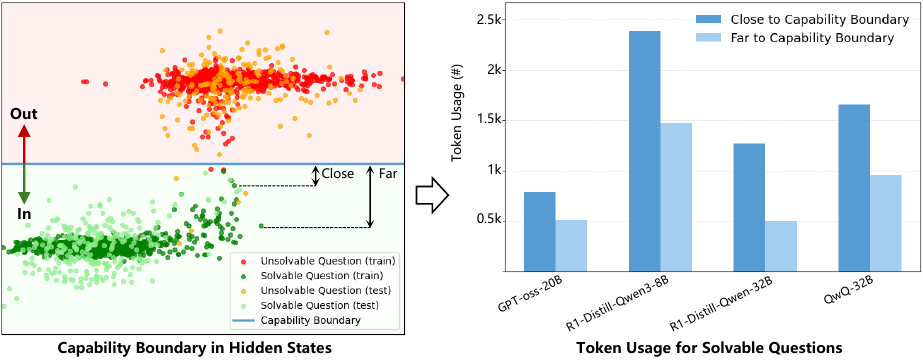}
    \vspace{-0.5em}
    \caption{\textbf{Left:} A linear classifier trained on hidden states clearly separates solvable and unsolvable questions under our setup. \textbf{Right:} Among solvable questions, those closer to the operational capability boundary require more token usage ($1.5$--$2\times$) to arrive at the correct answer.}
    \label{fig:CBhidden}
    \vspace{-1em}
\end{figure*}

To formalize these patterns, we design two indicators for determining whether a question is likely to lie within or beyond the capability boundary.

\noindent \textbf{Confidence Differential (ConfDiff)} measures the sign of gap between uncertain and confident trajectory along the reasoning process. Specifically, we accumulate the sign up to a given reasoning stage percentage $s\in[0,100]$, and treat the question as beyond the capability boundary if the cumulative value exceeds a threshold $\alpha_s$: 
\begin{equation}
\mathbb{P}_{t \in (0,s)}\!\left( \mathcal{D}_U(t) > \mathcal{D}_C(t) \right) > \alpha_s.
\label{eq:confdiff}
\end{equation}    

\noindent \textbf{Confidence Curvature (ConfCurv)} computes the second derivative of the expression density trajectory to capture its convexity \citep{boyd2004convex}. A concave trajectory corresponds to questions within the capability boundary, while a convex trajectory indicates questions beyond it. Also, we accumulate the sign of convexity up to a reasoning stage percentage $s$ and treat the question as beyond the capability boundary if the value exceeds a threshold $\beta_s$:
\begin{equation}
\mathbb{P}_{t \in (0,s)}\!\left( \frac{d^2}{dt^2}\,\big(\mathcal{D}_U(t)-\mathcal{D}_C(t)\big) < 0\right) > \beta_s.
\label{eq:confcurv}
\end{equation}    

\autoref{fig:confDiffCurv} shows that the two proposed indicators can reliably distinguish unsolvable from solvable questions, achieving high accuracy across different LRMs throughout the reasoning stage. Between the two indicators, ConfDiff is more reliable as it achieves higher accuracy and exhibits smaller fluctuations during the reasoning process. Moreover, strong predictive accuracy emerges early, e.g., as early as 2\% for GPT-oss-20B. Interestingly, the gradually declining yellow curve of this model suggests that these signals may be more salient at the beginning than at the end of reasoning. One possible explanation is that repetitive looping and error accumulation in later stages make the trajectories noisier \citep{yao2025understanding,mukherjee2025premise}. Overall, these results show that \textbf{\textit{reasoning expressions contain early signals related to the operational capability boundary and predictive of eventual failure}}.

\section{Capability Boundary in Hidden States}
\label{sec:cbHiddenStates}

Following the observation that reasoning expressions expose early signals related to the operational capability boundary, it is natural to ask whether similar signals may emerge even earlier, potentially before the model begins reasoning. We therefore investigate whether hidden states of the input question already contain information predictive of eventual success or failure.

Since hidden states are high-dimensional (e.g., 4096), limited samples may lead to sparsity issues that obscure statistical patterns \citep{vishwakarma2024curse,koppen2000curse}. We mix the mathematical benchmarks used in the previous sections and randomly split it into an 80\% training set and a 20\% test set, with balanced labels for eventually unsolved and solved questions.

Next, we extract hidden states from the prefilling stage as it processes the input question \citep{yuan2024llm,pope2023efficiently}, using the hook method \citep{vig2019multiscale}. Specifically, we focus on the hidden state of the last input token, which encapsulates the semantic content of the entire question \citep{dong2025ve,radford2019language}. Finally, we apply simple linear classifiers to these hidden states, including Linear Discriminant Analysis (LDA) and Logistic Regression (LR) \citep{hastie2009elements}. Using final answer correctness as labels, we test whether eventually solved and unsolved questions can be linearly separated, i.e., whether last token hidden states contain information predictive of eventual correctness under our setup.

\begin{table}
\centering
\resizebox{0.9\linewidth}{!}{
\begin{tabular}{l|c|ccc}
\toprule
                & \multirow{2}{*}{\textbf{LDA}} & \multicolumn{2}{c}{\textbf{LR}} \\
                &  & $C = 0.1$ & $C = 1$  \\
\midrule
GPT-oss-20B            & $98.0$  & $98.9$ &  $98.2$ \\
R1-Distill-Qwen3-8B    & $96.7$  & $96.7$  &  $97.0$ \\
R1-Distill-Qwen-32B    & $97.8$  & $97.8$ &   $98.0$  \\
QwQ-32B                & $98.9$  & $98.2$ &  $98.2$  \\
\bottomrule
\end{tabular}
}
\caption{Accuracy (\%) of separating solvable and unsolvable questions in hidden states using linear classifiers (robust across hyperparameter choices).}
\label{tab:accCBhidden}
\vspace{-2ex}
\end{table}

The left subfigure of \autoref{fig:CBhidden} visualizes the classification of solvable and unsolvable questions (see \autoref{app:CBhidden} for results of all LRMs), together with the decision boundary (blue line) of a linear classifier (e.g., LDA). We observe that the two classes of questions are clearly separated under random train–test splits. To quantify this separation, \autoref{tab:accCBhidden} reports the classification accuracy, which is consistently close to $100\%$ and robust to the choice of classifier and hyperparameters (e.g., the regularization coefficient $C$ of LR). These results suggest that the hidden states of the last input token contain information that is predictive of eventual correctness, and thus closely related to the operational capability boundary under our evaluation setup.

At the same time, these signals should be interpreted with care: they may partially reflect question difficulty rather than a purely intrinsic notion of capability boundary. We provide additional analyses in \autoref{app:whyHiddenStates}.

Interestingly, we also observe that among eventually solved questions, some lie very close to the classifier decision boundary, while others lie farther away. We quantify this distance by the probability difference assigned by the LDA classifier to the solved and unsolved classes \citep{hwang2023active}. We then compare these groups in terms of token usage during inference. Although LRMs answer both groups correctly, questions closer to the boundary require on average $1.5$--$2\times$ more tokens than those farther from the boundary (right subfigure of \autoref{fig:CBhidden}). This suggests that the hidden state signals are related not only to eventual correctness, but also to the amount of reasoning effort required to obtain a correct answer.

Overall, these results indicate that \textbf{\textit{the hidden states of the last input token contain useful signals related to the operational capability boundary, and that these signals also correlate with the reasoning effort required to solve a question.}}

Moreover, because these signals are extracted from reasoning expressions and hidden states rather than task-specific heuristics, they may extend beyond mathematical reasoning. We provide analyses on LiveCodeBench \citep{jain2024livecodebench} and GPQA-Diamond \citep{rein2023gpqa} in \autoref{app:coding}.

\section{Reason with Capability Boundary}
\label{sec:reasoningWithCB}

Because signals related to the operational capability boundary emerge early, they can be used to improve the efficiency and reliability of LRMs' responses. In this section, we propose two \textbf{\textit{test-time monitoring}} strategies that leverage such signals and demonstrate their effectiveness.

\subsection{Experimental Design}

\noindent \textbf{Strategies.} The key idea is to monitor signals of capability boundaries on the fly, either from reasoning trajectories (black-box) or hidden states (white-box), and to adjust reasoning when a question lies beyond the model's effective solving regime.

\noindent \textbf{Reason Expression Monitoring (Monitor\(_{\text{express}}\))} is a black-box strategy that leverages expressions density during reasoning. The pipeline is:  

\noindent \textbf{1. Construct expression trajectories:} continuously track the densities of confident and uncertain expressions (e.g., ``I think that's it'' vs. ``I might be wrong'') across reasoning steps to form expression density trajectories $\mathcal{D}_C(t)$ and $\mathcal{D}_U(t)$.  

\noindent \textbf{2. Predict likely-unsolved question:} analyze these expression trajectories on the fly using quantitative indicators such as Confidence Differential (ConfDiff) to determine whether a question is likely to remain unsolved.

\noindent \textbf{3. Early stop unproductive reasoning and reprompt with suffix:} if the trajectory indicates the question is likely to remain unsolved, we early terminate reasoning and reprompt the model to output a concise approach. This is done by appending a prompt suffix at the end of the question: 
\begin{tcolorbox}
    [width=\linewidth,colback={white},title={\fontsize{9.5}{7}\selectfont Prompt suffix },coltitle=white,left=1pt,right=1pt,top=1pt,bottom=1pt] 
{\small
The above question is beyond your capability boundary. Do not solve the question, just provide a concise potential approach of less than 5 steps:
}
\end{tcolorbox}

\noindent \textbf{Hidden States Monitoring (Monitor\(_{\text{hidden}}\))} is a white-box strategy that leverages the hidden states of the last input token. The pipeline is as follows:  
    
\noindent \textbf{1. Extract hidden states:} capture the hidden states of the last input token, which encodes the semantic representation of the entire question during the prefilling stage.

\noindent \textbf{2. Predict likely-unsolved question:} apply simple linear classifiers (e.g., LDA or LR) trained to distinguish eventually unsolved from solved questions. This prediction is made before reasoning begins.

\noindent \textbf{3. Avoid unproductive reasoning via constrained output prefix:} if the classifier predicts that the question is likely to remain unsolved under the current setup, the model abstains from full reasoning and instead outputs a concise possible approach. This is implemented by constraining the response to begin with an output prefix (see \autoref{app:ablationPrefix} for the ablation study):
\begin{tcolorbox}
    [width=\linewidth,colback={white},title={\fontsize{9.5}{7}\selectfont Output prefix },coltitle=white,left=1pt,right=1pt,top=1pt,bottom=1pt] 
{\small
\verb|<think>|

I think this question is beyond my capability boundary. I cannot fully solve it, but I can outline a concise potential approach. I must give a concise outline to the user (less than 10 steps)!

\verb|</think>|

This question is beyond my capability boundary, but I can outline a concise potential approach:

}
\end{tcolorbox}


\begin{table*}[t]
\centering
\resizebox{0.9\linewidth}{!}{
\begin{tabular}{l|cc|cc|cc}
\toprule
Context Length         & & & \multicolumn{2}{c|}{\textbf{2048}} & \multicolumn{2}{c}{\textbf{4096}} \\
Metric                        & \textbf{ACC (\%) ↑} & \textbf{HA (\%) ↑} & \textbf{Token ↓} & \textbf{Overflow (\%) ↓} & \textbf{Token ↓} & \textbf{Overflow (\%) ↓}  \\

\midrule
GPT-oss-20B  & $74.6$ &  $0$ & $2029$ &  $97.0$  & $4015$  &  $97.0$  \\
\hspace*{1em}+BoostAbstention & $74.6$\diffdown{0} &  $0$\diffup{0} & $2023$\diffdown{0.3} &  $97.0$\diffdown{0} & $3933$\diffdown{2.0}  &  $84.8$\diffdown{12.2}  \\
\hspace*{1em}+Monitor\(_{\text{express}}\) & $74.6$\diffdown{0}    &  $100$\diffup{100}   &  $1135$\diffdown{44.1}   &  $21.2$\diffdown{75.8}   &  $1499$\diffdown{62.7}     &   $15.2$\diffdown{81.8}       \\
\hspace*{1em}+Monitor\(_{\text{hidden}}\) & $73.7$\diffdown{0.9} & $98.9$\diffup{98.9} & $1784$\diffdown{12.1} & $78.4$\diffdown{18.6} & $3317$\diffdown{17.4}  &  $5.4$\diffdown{91.6}        \\
\midrule
R1-Distill-Qwen3-8B  & $76.9$ &  $0$ & $2048$ &  $100$  & $4096$  &  $100$  \\
\hspace*{1em}+BoostAbstention & $76.9$\diffdown{0}  &  $0$\diffup{0} & $2048$\diffdown{0} &  $100$\diffdown{0} & $4096$\diffdown{0}  &  $100$\diffdown{0}  \\
\hspace*{1em}+Monitor\(_{\text{express}}\) & $74.6$\diffdown{2.3}   &  $63.3$\diffup{63.3}   &  $1455$\diffdown{29.0}   &  $40.0$\diffdown{60.0}   &  $2551$\diffdown{37.7}     &   $30.0$\diffdown{70.0}       \\
\hspace*{1em}+Monitor\(_{\text{hidden}}\) & $73.8$\diffdown{3.1} & $97.9$\diffup{97.9} & $787$\diffdown{61.6} & $22.2$\diffdown{77.8} & $1243$\diffdown{69.7}  &  $0$\diffdown{100}        \\
\midrule
R1-Distill-Qwen-32B & $56.2$ & $0$  & $2048$ &  $100$  & $4095$ & $97.4$  \\
\hspace*{1em}+BoostAbstention & $56.2$\diffdown{0}  &  $0$\diffup{0} & $2005$\diffdown{2.1} &  $94.7$\diffdown{5.3}   & $3903$\diffdown{4.7}  &  $91.2$\diffdown{6.2}  \\
\hspace*{1em}+Monitor\(_{\text{express}}\)  &   $56.2$\diffdown{0}     &  $100$\diffup{100}    &  $801\diffdown{60.9}$  &  $0$\diffdown{100}   &  $801$\diffdown{80.4}     &    $0$\diffdown{97.4}      \\
\hspace*{1em}+Monitor\(_{\text{hidden}}\) & $55.3$\diffdown{0.9} & $97.0$\diffup{97.0} & $457$\diffdown{77.7} &  $5.3$\diffdown{94.7}  & $565$\diffdown{86.2} &  $5.3$\diffdown{92.1}   \\
\midrule
QwQ-32B         &  $71.5$  & $0$  &  $2048$ &   $100$  & $4096$ & $100$    \\
\hspace*{1em}+BoostAbstention & $71.5$\diffdown{0}  &  $0$\diffup{0} & $2048$\diffdown{0} &  $100$\diffdown{0}  & $4096$\diffdown{0}  &  $100$\diffdown{0}  \\
\hspace*{1em}+Monitor\(_{\text{express}}\)  &   $71.5$\diffdown{0}     &  $100$\diffup{100}   &  $1164$\diffdown{43.2}   & $32.4$\diffdown{67.6}    &  $1828$\diffdown{55.4}     &    $32.4$\diffdown{67.6}      \\
\hspace*{1em}+Monitor\(_{\text{hidden}}\) & $71.5$\diffdown{0}   & $96.9$\diffup{96.9}  & $262$\diffdown{87.2} & $0$\diffdown{100} & $262$\diffdown{93.6} & $0$\diffdown{100} \\
\bottomrule
\end{tabular}
}
\caption{Monitor\(_{\text{express}}\) and Monitor\(_{\text{hidden}}\) compared to baseline BoostAbstention. Our strategies enable LRMs to recognize questions beyond capability boundary (HA) without compromising accuracy (ACC), and provide more efficient response (Token usage and Overflow).
}
\label{tab:optimize}
\vspace{-1em}
\end{table*}

\noindent \textbf{Metrics.} We adopt four metrics to systematically evaluate the two proposed strategies: 
\begin{packeditemize}
    \item \textbf{Accuracy (ACC)} measures the proportion of correct answers, serving as the standard indicator of reasoning quality \citep{xia2025evaluating}.
    \item \textbf{Hard Abstention (HA)} measures the model’s ability to recognize questions beyond its capability boundary. Different to abstention which focus on underspecification, outdated information, or noncompliance \citep{kirichenko2025abstentionbench,madhusudhan2024llms,brahman2024art}, HA specifically evaluates whether LRMs can explicitly abstain on unsolvable questions instead of producing incorrect answers.
    \item \textbf{Token} usage measures the number of tokens consumed during reasoning and answering \citep{snell2024scaling}. This metric directly reflects the efficiency of the reasoning process.
    \item \textbf{Overflow} measures the proportion of cases where the model’s output exceeds the context window, which identifies incomplete or abnormal responses \citep{ben2025overflow}.  
\end{packeditemize}
Notably, ACC is evaluated on the full dataset, whereas the remaining three metrics are computed only on questions for which LRMs produce incorrect answers. For token usage and overflow, we vary the models' context length to observe how our strategies scale in longer context scenarios. \autoref{apptab:fullTradeoff} additionally reports false-positive abstention and token usage over all questions.

\noindent \textbf{Baseline.} Prior work has explored improving performance by self-verification after generating the answer \citep{madhusudhan2024llms, kadavath2022language} or by monitoring uncertainty and other undesirable behaviors \citep{baker2025monitoring,farquhar2024detecting,kuhn2023semantic}. However, these approaches are orthogonal to our scenario: the former introduces additional computation after producing an answer, while the latter does not address eventually unsolved questions (see more baseline comparison in \autoref{apptab:priorWork}). 

Therefore, we adopt a baseline that improve efficiency by enabling the model to abstain. In particular, we consider BoostAbstention \citep{kirichenko2025abstentionbench} which crafts a system prompt that encourages the model to abstain in designated scenarios, showing effectiveness for both reasoning and standard LLMs. For our experiments, we modify the system prompt to better match the abstention scenarios of eventually unsolved questions (see \autoref{app:boostAbstention} for the full prompt).

\subsection{Results}

\autoref{tab:optimize} compares the performance of the original models, the baseline BoostAbstention, and our proposed monitoring strategies. We highlight the following key observation.

\vspace{0.5em}
\noindent \textit{\textbf{Observation 1: our strategies enable LRMs to recognize eventually unsolved questions while maintaining comparable accuracy on solved ones.}} Although the original models achieve acceptable Accuracy (ACC), their Hard Abstention (HA) is zero, indicating a complete lack of ability to abstain on questions beyond the capability boundary. Moreover, almost all Overflow is 100\%, suggesting that most incorrect cases arise because LRMs struggling to solve a hard question until context limit. 

Promisingly, both strategies achieve strong performance. First, their ACC remains nearly unchanged compared to the original models, showing that good reasoning performance is preserved. Meanwhile, the substantially high HA (almost 100\%) demonstrates that LRMs can now effectively identify likely-unsolved questions. 

Surprisingly, and contrary to the observation from \citet{kirichenko2025abstentionbench}, the baseline BoostAbstention has almost no effect in our scenario. This reveals that mathematical reasoning problems differ from outdated or unsafe requests: abstention cannot be induced simply by system prompts, but instead requires boundary signals encoded in hidden states or reasoning expressions.

\begin{figure}
    \centering
    \includegraphics[width=0.99\linewidth]{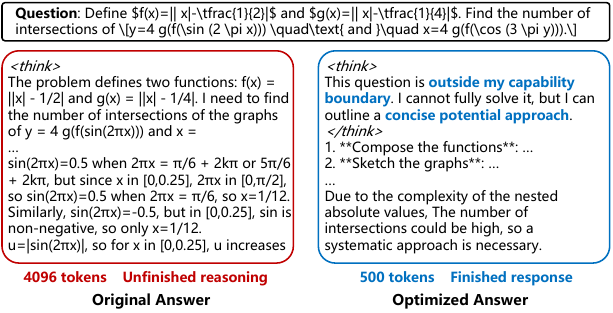}
    \caption{Our strategies avoid unproductive reasoning by acknowledging eventually unsolved questions, improving efficiency and reliability of response.}
    \label{fig:optimizedAnswer}
    \vspace{-1.5em}
\end{figure}

\vspace{0.5em}
\noindent \textit{\textbf{Observation 2: our strategies enable LRMs to provide more efficient and reliable responses to eventually unsolved questions by offering concise solution approaches.}} \autoref{fig:optimizedAnswer} illustrates the responses produced by our strategies compared to the original models. The original models struggle with eventually unsolved questions and fail to complete the reasoning process even after exhausting the context length. Such responses provide neither a correct answer nor valuable insights from the lengthy reasoning trace, while also wasting time. 

In contrast, our strategies detect these unsolved questions and guide the model to explicitly acknowledge that they are beyond capability boundary, while offering a concise possible approach. This response paradigm provides users with possible hints for problem solving without wasting time on meaningless output. And importantly, it does not compromise the model’s ability to produce correct answers for solvable questions. \autoref{apptab:outputUtility} provides an additional utility evaluation of these concise outputs.

This explains why we observe large reductions in both token usage (up to 93.6\%) and overflow (up to 100\%) in \autoref{tab:optimize}, indicating that reasoning with capability boundaries greatly improves efficiency and mitigates incomplete outputs. Besides, this reduction becomes more evident with larger context length. For example, overflow reduces from 78.4\% to 5.4\% on GPT-oss-20B when the context length is extended from 2048 to 4096. This suggests that our strategies hold promise for optimizing long-context reasoning tasks \citep{ling2025longreason,kuratov2024babilong}. We provide in \autoref{app:maximumContextLength} an additional analysis with the context length set to maximum.

\vspace{0.5em}
\noindent \textit{\textbf{Observation 3: LRMs vary in reasoning behavior correction, with some good at instruction following and others more prone to output guidance.}} We compare the performance of Monitor\(_{\text{express}}\) and Monitor\(_{\text{hidden}}\) across different LRMs. For GPT-oss-20B, Monitor\(_{\text{express}}\) achieves lower token usage than Monitor\(_{\text{hidden}}\). We find this model is resistant to output guidance via constrained output prefix: although it sometimes begins by generating a solution approach, it tends to later negate this and restart the reasoning process, resulting in long and unproductive outputs. In contrast, its instruction-following ability is relatively strong, and thus reprompting with the prompt suffix via Monitor\(_{\text{express}}\) yields lower token usage.

Conversely, the other three LRMs show weaker ability of instruction following  but more prone to output guidance, making Monitor\(_{\text{hidden}}\) more effective than Monitor\(_{\text{express}}\). This suggests that, in addition to the black-box versus white-box distinction, the choice of optimization strategy in practice should also consider LRMs' behavioral tendencies.

\section{Conclusion}
We study whether signals related to the operational capability boundary can help mitigate unproductive reasoning in LRMs. Across both black-box and white-box analyses, we find that LRMs expose such signals early: reasoning expressions differ systematically between eventually solved and unsolved questions, while the hidden states of the last input token are predictive of eventual correctness even before reasoning begins. Based on these findings, we propose two test-time monitoring strategies that reduce unproductive reasoning. Experiments show that they largely preserve accuracy on solvable questions while substantially improving efficiency and reliability on questions lie beyond the operational capability boundary.

\clearpage
\section*{Limitations}
\noindent \textbf{Interpretation of capability boundaries.} In this paper, we use \emph{capability boundary} in an operational sense, namely as the empirical separation between questions a model eventually solves and those it does not under a fixed evaluation setup. Accordingly, the signals we identify should be interpreted as predictive correlates of eventual failure rather than definitive evidence of a formally defined intrinsic capability boundary. In particular, these signals may partially reflect question difficulty. Although we attempt to mitigate this concern through per-benchmark and cross-benchmark analyses (\autoref{app:whyHiddenStates}), the extent to which these signals capture a general notion of capability boundary remains an open question. Developing a more principled formalization of capability boundaries (e.g., disentangling question difficulty) is an important direction for future work.

\noindent \textbf{Scope of tasks and evaluation domains.} Our evaluation focuses primarily on mathematical benchmarks, which allow systematic analysis of unsolvability signals but may not fully represent other reasoning scenarios. While our experiments on LiveCodeBench \citep{jain2024livecodebench} and GPQA-Diamond \citep{rein2023gpqa} suggest similar results (\autoref{app:coding}), the generality of capability boundary signals across broader domains (e.g., complex decision making or scientific reasoning) remains to be established. Future work should investigate whether similar strategies apply across more diverse tasks.

\noindent \textbf{Access assumptions.} Monitor\(_{\text{express}}\) requires complete, visible reasoning traces, while Monitor\(_{\text{hidden}}\) requires access to the last-token hidden state. Therefore, the former cannot be directly applied when chain-of-thought is hidden, truncated, or summarized, and the latter cannot be applied to closed-source APIs unless hidden states are exposed.

\section*{Ethics Statement}
ACL Ethics Policy is respected in this work. It studies LRMs' capability boundaries, aiming to improve efficiency and reliability. No human subjects, sensitive personal data, or harmful content were involved. The proposed methods are intended for research purposes only and do not pose foreseeable risks of misuse.

\section*{Acknowledgements}
This work was supported by Ant Group.

\section*{Use of AI Assistants}
The authors used AI assistants for language polishing, LaTeX editing, and phrasing suggestions during paper preparation. All substantive claims, experimental results, analyses, citations, and final text were reviewed and verified by the authors.

\bibliography{custom}

\clearpage
\appendix

\section{Datasets and Models}
\label{app:modelsAndDataset}

In this section, we provide the specific description of models and datasets evaluated in our work. 

\autoref{apptab:evaluationScope} summarizes the evaluation scope in the original submission. The approximately 120 questions refer only to the four competition-math benchmarks, rather than to the full evaluation.

\begin{table}[H]
\centering
\small
\setlength{\tabcolsep}{4pt}
\begin{tabular}{@{}p{0.4\columnwidth}p{0.32\columnwidth}r@{}}
\toprule
\textbf{Datasets} & \textbf{Task type} & \textbf{Size} \\
\midrule
AIME'25 / AIME'24 / HMMT / AMC'23 & Competition math & 120 \\
GSM8K & Grade-school math & 1.5k \\
HLE & Broad expert QA & 1.5k \\
LiveCodeBench & Coding & 1055 \\
\bottomrule
\end{tabular}
\caption{Evaluation scope in the original submission. HLE uses text-only questions.}
\label{apptab:evaluationScope}
\end{table}

We evaluate the following LRMs:
\begin{packeditemize}
    \item \textbf{GPT-oss-20B} \citep{agarwal2025gpt}: An open-source large reasoning model with 20B parameters, designed as a general-purpose backbone for multi-step reasoning tasks.
    \item \textbf{DeepSeek-R1-0528-Qwen3-8B} \citep{guo2025deepseek}: A distilled variant of DeepSeek-R1 based on Qwen3-8B, post-trained with chain-of-thought supervision. It achieves state-of-the-art performance among open-source models on mathematical reasoning benchmarks.
    \item \textbf{DeepSeek-R1-Distill-Qwen-32B} \citep{guo2025deepseek}: A 32B-parameter distilled model obtained by transferring reasoning abilities from DeepSeek-R1 into the Qwen2.5-32B backbone, yielding strong performance across benchmarks.
    \item \textbf{QwQ-32B} \citep{qwq-32b}: A medium-sized reasoning model from the Qwen series, competitive with other state-of-the-art reasoning-focused models on math and logical reasoning tasks.
\end{packeditemize}
For the sampling settings, we use greedy decoding to ensure reproducibility of results. For the maximum sequence length, we follow the specifications in the HuggingFace model cards: 128k for GPT-oss-20B, 64k for DeepSeek-R1-0528-Qwen3-8B, and 32k for both DeepSeek-R1-Distill-Qwen-32B and QwQ-32B.


We evaluate the following Math datasets:
\begin{packeditemize}
\item \textbf{AIME'25} \citep{aime2025}: The 2025 American Invitational Mathematics Examination. It consists of 30 integer-answer questions with increasing difficulty, used as a challenging benchmark for mathematical reasoning.

\item \textbf{AIME'24} \citep{aime2024}: The 2024 American Invitational Mathematics Examination. Similar in format to AIME'25, it consists of 30 problems and provides a prior-year benchmark for evaluating model robustness across editions.

\item \textbf{HMMT} \citep{hmmt_feb_2025}: The Harvard–MIT Mathematics Tournament (February 2025 session). The dataset contains 30 questions sourced from the original competition, which were extracted, converted to LaTeX, and verified. These problems feature complex multi-step reasoning across various mathematical domains.

\item \textbf{AMC'23} \citep{amc2023_10a}: The 2023 American Mathematics Competition (AMC 10A). The dataset consists of 30 questions, providing a lighter but diverse benchmark compared to AIME and HMMT. 

\item \textbf{GSM8K} \citep{cobbe2021training}: A large-scale dataset of grade-school math word problems requiring multi-step arithmetic reasoning. We evaluate on approximately 1.5k problems from the test set. It is widely used to evaluate the basic reasoning competence of LRMs.

\item \textbf{Humanity's Last Exam (HLE)} \citep{phan2025humanity}: A multi-modal benchmark at the frontier of human knowledge, designed with broad subject coverage across mathematics, humanities, and natural sciences. We use approximately 1.5k text-only questions (excluding image-based problems) from this dataset. 
\end{packeditemize}

The evaluated benchmarks also differ substantially in question length, mathematical notation, answer-format requirements, and prompt openings. \autoref{apptab:promptStyle} reports these surface-style indicators, showing that Monitor\(_{\text{express}}\) is not evaluated on a single fixed question template.

\begin{table*}[t]
\centering
\small
\setlength{\tabcolsep}{7pt}
\begin{tabular}{@{}lrrrrr@{}}
\toprule
\textbf{Benchmark} & \textbf{Avg. words} & \textbf{Math spans/Q} & \textbf{Math-char (\%)} & \textbf{Answer-transform (\%)} & \textbf{Direct-start (\%)} \\
\midrule
AIME'24       & 63.6  & 6.8  & 19.0 & 30.0 & 13.3 \\
AIME'25       & 87.2  & 10.2 & 21.8 & 31.0 & 10.3 \\
AMC'23        & 53.7  & 5.7  & 20.7 & 30.8 & 23.1 \\
HMMT Feb. 2025 & 63.7 & 6.5  & 26.2 & 0.0  & 17.2 \\
GSM8K         & 46.8  & 0.2  & 5.2  & 62.9 & 0.1 \\
HLE text-only & 111.3 & 7.5  & 28.6 & 80.6 & 11.6 \\
\bottomrule
\end{tabular}
\caption{Prompt-style diversity in the evaluated benchmarks.}
\label{apptab:promptStyle}
\end{table*}



\section{Limitations of Some Prior Work}
\label{app:limitationsPriorWork}

Some prior works evaluate model confidence through self-verification \citep{kadavath2022language} or uncertainty estimation \citep{farquhar2024detecting, kuhn2023semantic}. In this section, we show that these approaches are less effective than our proposed monitoring strategies for identifying unsolvable questions with comparable accuracy on solvable ones. We first clarify how we adapt these methods to fit our task.

\noindent \textbf{Self-verification (SV).} For each question, we ask the model to evaluate whether its own predicted answer is correct using the prompt template:

\begin{tcolorbox}
    [width=\linewidth,colback={white},title={\fontsize{9.5}{7}\selectfont Self-verification prompt },coltitle=white,left=1pt,right=1pt,top=1pt,bottom=1pt] 
{\small
Question: \{question\}

Proposed Answer: \{answer\}

Is the proposed answer:

A True

B False

The proposed answer is:

}
\end{tcolorbox}

This follows prior work \citep{kadavath2022language} which study whether language models can evaluate the validity of their own claims by asking models to first propose answers, and then to evaluate that their answers are correct. If the model judges its own incorrect answer to be incorrect, we count this as a hard abstention (HA). Conversely, if the model judges its correct answer to be incorrect, this reduces the overall task accuracy (Acc). Formally,
\begin{equation}
\begin{aligned}
\text{Acc} =
\mathbb{P}(&\text{Model judges the answer to be correct} \\
&\land \text{Answer is correct})
\end{aligned}
\end{equation}
\begin{equation}
\begin{aligned}
\text{HA} =
\mathbb{P}(&\text{Model judges the answer to be incorrect} \\
&\mid \text{Answer is incorrect})
\end{aligned}
\end{equation}

\noindent \textbf{Uncertainty-estimation (UE).} For each question, we compute the semantic entropy of the model's generated answer, where a higher entropy indicates greater uncertainty. This follows prior work \citep{farquhar2024detecting, kuhn2023semantic}, which introduces semantic entropy as an uncertainty measure that captures linguistic invariances induced by shared meanings. When the entropy exceeds a certain threshold, the model is considered uncertain about its answer. Similar as before, we define:
\begin{equation}
\begin{aligned}
\text{Acc} =
\mathbb{P}(&\text{Semantic entropy of answer} < \text{threshold} \\
&\land \text{Answer is correct})
\end{aligned}
\end{equation}
\begin{equation}
\begin{aligned}
\text{HA} =
\mathbb{P}(&\text{Semantic entropy of answer} \ge \text{threshold} \\
&\mid \text{Answer is incorrect})
\end{aligned}
\end{equation}

We additionally compare with BudgetPrompt \citep{han2025token}, which constrains the reasoning budget through prompting, and Dynamic-Early-Exit \citep{yang2025dynamic}, which stops reasoning dynamically.

\begin{table}[t]
\centering
\small
\setlength{\tabcolsep}{5pt}
\begin{tabular}{@{}lrrr@{}}
\toprule
\textbf{Method} & \textbf{ACC $\uparrow$} & \textbf{HA $\uparrow$} & \textbf{FPA $\downarrow$} \\
\midrule
GPT-oss-20B & 74.6 & 0.0 & 0.0 \\
\hspace*{0.7em}+BoostAbstention & 74.6 & 0.0 & 0.0 \\
\hspace*{0.7em}+BudgetPrompt & 46.7 & 0.0 & 37.4 \\
\hspace*{0.7em}+Self-verification & 32.1 & 46.0 & 57.0 \\
\hspace*{0.7em}+Uncertainty estimation & 50.0 & 29.0 & 33.0 \\
\hspace*{0.7em}+Dynamic-Early-Exit & 49.8 & 0.0 & 33.2 \\
\hspace*{0.7em}+Monitor$_{\text{express}}$ & \textbf{74.6} & \textbf{100.0} & \textbf{0.0} \\
\hspace*{0.7em}+Monitor$_{\text{hidden}}$ & 73.7 & 98.9 & 1.2 \\
\bottomrule
\end{tabular}
\caption{Comparison to five baselines on GPT-oss-20B (\%). FPA denotes false-positive abstention, $P(\text{abstain}\mid\text{originally correct})$.}
\label{apptab:priorWork}
\end{table}

\noindent \textbf{Results.} \autoref{apptab:priorWork} shows that BoostAbstention, BudgetPrompt, and Dynamic-Early-Exit do not identify unsolvable questions (HA = 0). Self-verification and uncertainty estimation provide some HA only while sacrificing many originally correct cases. In contrast, our monitors preserve ACC while achieving high HA and low FPA.

\section{Confident and Uncertain Expressions}
\label{app:confidentUncertainExpression}

\begin{figure*}[t]
    \centering
    \includegraphics[width=1\linewidth]{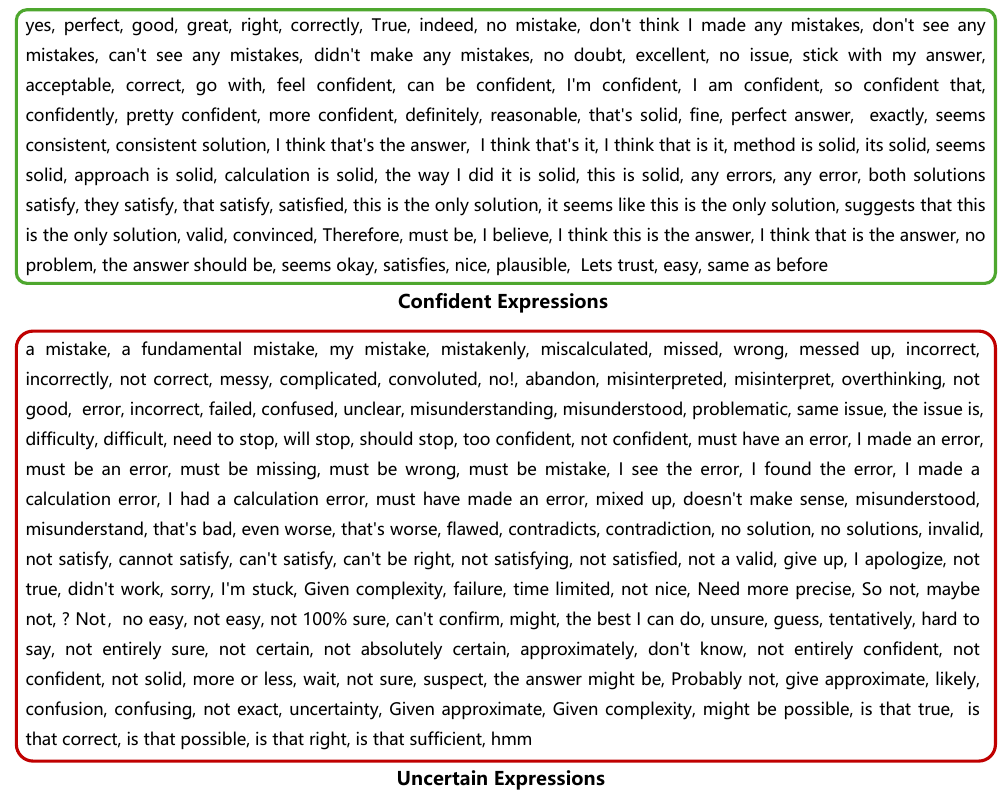}
    \caption{Confident vs. uncertain expressions.}
    \label{appfig:listConfUncertainExpression}
\end{figure*}

To ensure that our analysis is not tied to particular LRM's reasoning style, we extract confident and uncertain expressions individually for each model, based on its own characteristic reasoning patterns. The extraction pipeline is outlined below.

\textbf{Step 1: Identify all anthropomorphic expressions from the model’s reasoning trace.} Prior work \citep{guo2025deepseek,qian2025demystifying} show that the anthropomorphic expressions (e.g., ``wait'') are crucial for LRM reasoning performance, in contrast to purely neutral mathematical steps (e.g., ``11 + 5 = 16''). Following this insight, we curate a set of anthropomorphic expressions that the LRM commonly uses. And we further verify with GPT-5 that these expressions indeed convey anthropomorphic tone rather than purely propositional content.

\textbf{Step 2: Select expressions whose occurrence frequency is most predictive of answer correctness.} For each question, we count the occurrences of all extracted expressions within reasoning. Using the final answer correctness as the label and the occurrence counts for each expressions as features, we apply feature-importance analysis methods \citep{wang2024feature}, including SHAP and Permutation Importance (PI), to identify the subset of expressions most informative for predicting whether the model's final answer is correct. As shown in \autoref{tab:featureImportanceScore} (an example for GPT-oss-20B), we select expressions such as ``wait'', ``yes'', and ``exactly'' which rank among the most predictive, whereas others (e.g., ``so'') exhibit negligible predictive value and are therefore excluded.

\begin{table}[H]
\centering
\small
\resizebox{0.8\linewidth}{!}{
\begin{tabular}{l|cc|c}
\toprule
\textbf{Expression} & \textbf{SHAP} $\uparrow$ & \textbf{PI} $\uparrow$ & \textbf{Total} $\uparrow$ \\
\midrule
wait     & 1.00 & 1.00 & 2.00 \\
yes      & 0.33 & 0.73 & 1.06 \\
exactly  & 0.26 & 0.73 & 0.99 \\
\ldots   &      &      &      \\
so       & 0.01 & 0.00 & 0.01 \\
\bottomrule
\end{tabular}
}
\vspace{-1ex}
\caption{
Feature importance score of expressions (shown here for GPT-oss-20B).
}
\label{tab:featureImportanceScore}
\vspace{-2ex}
\end{table}

\textbf{Step 3: Divide the selected expressions into confidence and uncertainty categories.} This step is grounded in the connection between epistemic markers and expressed uncertainty. Prior work \citep{liu2025revisiting,yona2024can} show that epistemic markers reflect LLMs' internal confidence, analogous to how humans use expressions such as ``I am not sure'' or ``it is unlikely that'' to convey uncertainty. Following this insight, we classify each selected expression as reflecting confidence or uncertainty (further validated by GPT-5).

\autoref{appfig:listConfUncertainExpression} shows the full list of expressions of all LRMs. When matching these expressions, the \textit{Uncertain Expression} is matched first, followed by the \textit{Confident Expression}. Additionally, both the position of the first token of an expression and the token length of the entire expression are considered.

We estimate the model’s confidence level from the occurrences of confidence and uncertainty expressions throughout the reasoning process, following prior work \citep{liu2025revisiting} which measures LLM uncertainty by counting the frequency of epistemic markers. In \autoref{sec:cannotSolve}, we use the category of expressions that appears more frequently near the end of the reasoning trace to determine whether the model is more confident or more uncertain. Building on this idea, \autoref{sec:cbReasoningTokens} analyzes the trajectory of confidence and uncertainty signals over the entire reasoning sequence, captured by ConfDiff (\autoref{eq:confdiff}) and ConfCurv (\autoref{eq:confcurv}), to assess whether the model is more confident or more uncertain.

\paragraph{Phrase-list sensitivity.}
We compare the full phrase-construction pipeline with three alternatives: removing generic expressions, retaining only top-ranked expressions, and randomly dropping half of the expressions. For each model, we tune the operating point using the full expression signal and apply the same operating point to the ablated phrase sets. \autoref{apptab:phraseSensitivity} shows that the full pipeline performs best on average, while the ablated lists remain predictive but weaker.

\begin{table}[H]
\centering
\small
\resizebox{0.99\linewidth}{!}{
\begin{tabular}{l c c c c}
\toprule
\textbf{Model} & \makecell{\textbf{Remove}\\\textbf{generic}} & \makecell{\textbf{Top-}\\\textbf{ranked}} & \makecell{\textbf{Random}\\\textbf{half}} & \textbf{Ours} \\
\midrule
GPT-oss-20B & 81.1 & 85.0 & 80.6 & \textbf{85.0} \\
R1-Distill-Qwen3-8B & 75.4 & 71.5 & 51.1 & \textbf{82.3} \\
R1-Distill-Qwen-32B & 75.4 & 83.1 & 71.8 & \textbf{89.2} \\
QwQ-32B & 66.9 & 78.5 & 65.4 & \textbf{83.8} \\
\midrule
Average & 74.7 & 79.5 & 67.2 & \textbf{85.1} \\
\bottomrule
\end{tabular}
}
\caption{Phrase-list sensitivity across models (separating accuracy, \%).}
\label{apptab:phraseSensitivity}
\end{table}

\paragraph{Threshold tuning.}
We use a strict train/validation/test split: the training split constructs the expression signals, the validation split tunes the stage-wise thresholds, and the test split is used only for final evaluation. For each stage $s$, ConfDiff is computed as
\begin{equation}
R^{\mathrm{diff}}_{i,s}=\frac{1}{s}\sum_{t=1}^{s}\mathbf{1}\!\left[\mathcal{D}_U^i(t)>\mathcal{D}_C^i(t)\right],
\end{equation}
and $\alpha_s$ is selected on the validation split by
\begin{equation}
\begin{aligned}
\alpha_s^*=\arg\max_{\alpha}\frac{1}{2}\big[&
\Pr(R^{\mathrm{diff}}_{i,s}>\alpha\mid y_i=1)\\
&+\Pr(R^{\mathrm{diff}}_{i,s}\leq\alpha\mid y_i=0)\big],
\end{aligned}
\end{equation}
where $y_i=1$ means that the original answer is incorrect. Likewise,
\begin{equation}
R^{\mathrm{curv}}_{i,s}=\frac{1}{s-1}\sum_{t=1}^{s-1}\mathbf{1}\!\left[\Delta^2(\mathcal{D}_U^i-\mathcal{D}_C^i)(t)<0\right],
\end{equation}
and $\beta_s$ is selected by the same balanced-accuracy criterion:
\begin{equation}
\begin{aligned}
\beta_s^*=\arg\max_{\beta}\frac{1}{2}\big[&
\Pr(R^{\mathrm{curv}}_{i,s}>\beta\mid y_i=1)\\
&+\Pr(R^{\mathrm{curv}}_{i,s}\leq\beta\mid y_i=0)\big].
\end{aligned}
\end{equation}
The selected thresholds are frozen before evaluating the test split.

\autoref{apptab:strictSplit} reports a strict-split diagnostic. For ConfCurv and ConfDiff, expression lists are built on train and thresholds are tuned on validation. For hidden states, LR and LDA are trained on train, with LR regularization tuned on validation. The strict-split accuracy scores remain close to the original diagnostics.

\begin{table}[H]
\centering
\small
\resizebox{0.99\linewidth}{!}{
\begin{tabular}{l c c}
\toprule
\textbf{Signal/classifier} & \textbf{Original Acc.} & \textbf{Strict-split Acc.} \\
\midrule
ConfCurv & 86.9 & 85.1 \\
ConfDiff & 86.2 & 86.2 \\
LR       & 98.0 & 97.8 \\
LDA      & 97.9 & 97.8 \\
\bottomrule
\end{tabular}
}
\caption{Strict train/validation/test split diagnostic (accuracy, \%).}
\label{apptab:strictSplit}
\end{table}


\section{System Prompt of BoostAbstention}
\label{app:boostAbstention}

We consider BoostAbstention \citep{kirichenko2025abstentionbench} which crafts a system prompt that encourages the model to abstain in designated scenarios, showing effectiveness for both reasoning and standard LLMs. For our experiments, we modify the system prompt to better match the abstention scenarios of questions beyond capability boundary:

\begin{tcolorbox}
    [width=\linewidth,colback={white},title={\fontsize{9.5}{7}\selectfont System Prompt of BoostAbstention},coltitle=white,left=1pt,right=1pt,top=1pt,bottom=1pt] 
{\small
You are a helpful assistant specialized in solving math problems. 
\\
\\
However, some problems may be too complex or beyond your capability. In such cases:\\
- Admit that you cannot fully solve the problem.\\
- Provide a concise potential approach, outline, or next step instead of a detailed solution.\\  
- Avoid spending too much time on overly difficult problems, as this increases latency for the user.  
}
\end{tcolorbox}


\section{Ablation Study on Output Prefix}
\label{app:ablationPrefix}

During our initial exploration, we experiment with several output prefixes and select the one that achieves the largest reduction in token usage. We conduct an ablation study comparing:
\begin{packeditemize}
    \item \textbf{Initial prefix (our early attempt)}: Our goal is to prompt the model to acknowledge that it cannot solve an unsolvable question and then provide only a concise potential approach. However, we unexpectedly found that this cannot minimize token usage. The model’s instruction-following ability was insufficient for reliably producing only a concise approach, it often generated overly long responses. The initial prefix is shown below:
\begin{tcolorbox}
    [width=\linewidth,colback={white},title={\fontsize{9.5}{7}\selectfont Initial output prefix },coltitle=white,left=1pt,right=1pt,top=1pt,bottom=1pt] 
{\small
\verb|<think>|

I think this question is beyond my capability boundary. I cannot fully solve it, but I can outline a concise potential approach.

\verb|</think>|

This question is beyond my capability boundary, but I can outline a concise potential approach:
}
\end{tcolorbox}

    \item \textbf{Emphasis prefix (final choice)}: Inspired by prior work \citep{zhang2024understanding,zhao2021calibrate} which show that the final sentence in the instruction strongly influences model behavior, we appended \texttt{Step 1:} at the end of the instruction to explicitly trigger step-by-step approach. Also, we emphasize in the end of the \texttt{<think>...</think>} part that \texttt{I must give a concise outline to the user (less than 10 steps)!}. This combination elicits short, structured responses and lead to the largest reduction in token usage. This is the output prefix we ultimately choose:
\begin{tcolorbox}
    [width=\linewidth,colback={white},title={\fontsize{9.5}{7}\selectfont Emphasis output prefix },coltitle=white,left=1pt,right=1pt,top=1pt,bottom=1pt] 
{\small
\verb|<think>|

I think this question is beyond my capability boundary. I cannot fully solve it, but I can outline a concise potential approach. I must give a concise outline to the user (less than 10 steps)!

\verb|</think>|

This question is beyond my capability boundary, but I can outline a concise potential approach:

Step 1:
}
\end{tcolorbox}

    \item \textbf{Token-budget prefix (following prior work)}: Following prior work \citep{han2025token} which compresses unnecessary reasoning by including a reasonable token budget in the prompt, we replace \texttt{less than 10 steps} with an explicit budget constraint \texttt{use less than 500 tokens}. The token-budget prefix is:
\begin{tcolorbox}
    [width=\linewidth,colback={white},title={\fontsize{9.5}{7}\selectfont Token-budget output prefix },coltitle=white,left=1pt,right=1pt,top=1pt,bottom=1pt] 
{\small
\verb|<think>|

I think this question is beyond my capability boundary. I cannot fully solve it, but I can outline a concise potential approach. I must give a concise outline to the user (less than 500 tokens)!

\verb|</think>|

This question is beyond my capability boundary, but I can outline a concise potential approach:

Step 1:
}
\end{tcolorbox}

\end{packeditemize}

To avoid the confounding effect of incomplete answers, we set the context length to the maximum according to Huggingface model card.

\autoref{apptab:ablationPrefix} shows that \textit{all three designs reduce token usage, with the Emphasis prefix achieving the largest reduction}. While it is possible that an even better prompt design exists, this does not affect our core conclusion: \textit{setting an output prefix enables the model to provide more efficient and more reliable responses on unsolvable questions}.

\begin{table}[H]
\centering
\small
\resizebox{0.6\linewidth}{!}{
\begin{tabular}{lc}
\toprule
\textbf{Output prefix} & \textbf{Token} $\downarrow$ \\
\midrule
No output prefix    & 5096 \\
Initial             & 4423 \\
\textbf{Emphasis}   & \textbf{2227} \\
Token-budget        & 2567 \\
\bottomrule
\end{tabular}
}
\vspace{-1ex}
\caption{
Token usage under different output prefixes.
}
\label{apptab:ablationPrefix}
\end{table}

\section{Additional Analysis of Capability Boundary Signals in Hidden States}
\label{app:whyHiddenStates}

In \autoref{sec:cbHiddenStates}, we show that eventually solved and unsolved questions can be separated using signals extracted from the hidden states of the last input token. However, these signals may partially correlate with question difficulty rather than reflecting a purely intrinsic notion of capability boundary.

In this section, we do not aim to establish a clean separation between capability boundary signals and question difficulty. Instead, we provide additional analyses showing that the hidden state signals identified in our setup remain informative under stricter per-benchmark and cross-benchmark settings.

We also compare the hidden-state classifier with controls based on benchmark/source identity and text difficulty proxies, including surface features such as length, format, and notation density. \autoref{apptab:controlClassifiers} shows that these controls carry some signal, but are substantially weaker than the last-token hidden states. Together with the held-out results in \autoref{apptab:strictSplit}, this supports the narrower interpretation that the hidden states provide predictive information beyond the tested metadata and text proxies.

\begin{table}[H]
\centering
\small
\resizebox{0.99\linewidth}{!}{
\begin{tabular}{l c c c}
\toprule
\textbf{Model} & \makecell{\textbf{Benchmark/}\\\textbf{source}} & \makecell{\textbf{Text difficulty}\\\textbf{proxies}} & \makecell{\textbf{Last-token}\\\textbf{hidden states}} \\
\midrule
GPT-oss-20B & 63.4 & 55.6 & \textbf{98.0} \\
R1-Distill-Qwen3-8B & 66.7 & 61.1 & \textbf{96.7} \\
R1-Distill-Qwen-32B & 63.6 & 60.9 & \textbf{97.8} \\
QwQ-32B & 66.5 & 71.6 & \textbf{98.9} \\
\bottomrule
\end{tabular}
}
\caption{Control classifier comparison across models (separating accuracy, \%).}
\label{apptab:controlClassifiers}
\end{table}

\noindent \textbf{An intuition from human exams.} One possible intuition comes from human test taking. In an exam with limited time and many difficult problems, skilled test-takers may quickly identify questions that are unlikely to be solved within the available budget and avoid wasting excessive time on them. This judgment may come either from recognizing that a question resembles problems they have repeatedly struggled with in the past (analogous to our Monitor$_{\text{hidden}}$ strategy), or from taking a few initial steps and realizing that they lack confidence to proceed (analogous to our Monitor$_{\text{express}}$ strategy). We present this analogy only as intuition, not as evidence of human-like metacognitive awareness in LRMs. Motivated by this intuition, we further conduct per-benchmark and cross-benchmark analyses as follows.

\begin{table}[H]
\centering
\small
\resizebox{0.8\linewidth}{!}{
\begin{tabular}{l| l| c c}
\toprule
\textbf{Dataset} & \textbf{Model} & \textbf{LDA} & \textbf{LR} \\
\midrule
\multirow{4}{*}{HLE}     & GPT-oss-20B          & 89.3 & 90.3 \\
    & R1-Distill-Qwen3-8B  & 94.9 & 96.4 \\
    & R1-Distill-Qwen-32B & 96.4 & 96.4 \\
    & QwQ-32B              & 93.4 & 96.4 \\
\midrule
\multirow{3}{*}{GSM8k}   & GPT-oss-20B          & 80.3 & 83.3 \\
  & R1-Distill-Qwen3-8B  & 70.1 & 71.6 \\
  & R1-Distill-Qwen-32B & 85.2 & 89.0 \\
  & QwQ-32B              & 87.9 & 87.1 \\
\bottomrule
\end{tabular}
}
\caption{
Accuracy (\%) of separating solvable and unsolvable questions using linear classifiers trained and tested within the same benchmark.
}

\label{apptab:perBenchmark}
\end{table}

\noindent \textbf{Per-benchmark test.} We provide in \autoref{apptab:perBenchmark} the accuracy of separating unsolvable questions from solvable ones \textit{using LDA and LR classifiers trained and tested on splits from the same dataset}. Specifically, for HLE and GSM8K, we randomly split each dataset into an 80\% training set and a 20\% test set (with balanced solvable/unsolvable labels), train the linear classifier on the training set, and evaluate its classification accuracy on the test set.

The results show that \textit{linear classifiers can still separate solvable and unsolvable questions from hidden states, although with reduced accuracy compared to aggregated benchmarks setting (\autoref{tab:accCBhidden})}. This drop is small on HLE but larger on GSM8K. The reason is that \textit{HLE contains a larger proportion of unsolvable questions, enabling the linear classifiers to better learn the hidden state information associated with unsolvability}. Consequently, these unsolvable questions in HLE can be effectively leveraged to develop models' Hard Abstention (HA) capability.

\noindent \textbf{Cross-benchmark test.} In the above analysis, we think the unsolvable questions in HLE can be effectively leveraged to develop models' Hard Abstention (HA) capability. Therefore, we train linear classifiers on HLE and evaluate models' Hard Abstention capability on GSM8k and AIME to test whether this capability generalizes across benchmarks. As shown in \autoref{apptab:crossBenchmark}, linear classifiers trained on HLE are able to identify nearly all unsolvable questions in GSM8K and AIME, achieving HA close to 100\%.

\begin{table}[H]
\centering
\small
\resizebox{0.7\linewidth}{!}{
\begin{tabular}{l|cc}
\toprule
\textbf{Model} & \textbf{GSM8k} & \textbf{AIME} \\
\midrule
GPT-oss-20B                & 0     & 0     \\
\hspace*{1em}+Monitor$_{\text{hidden}}$ & 100   & 100   \\
\midrule
R1-Distill-Qwen3-8B        & 0     & 0     \\
\hspace*{1em}+Monitor$_{\text{hidden}}$ & 100   & 100   \\
\midrule
R1-Distill-Qwen-32B       & 0     & 0     \\
\hspace*{1em}+Monitor$_{\text{hidden}}$ & 90.5  & 94.7  \\
\midrule
QwQ-32B                    & 0     & 0     \\
\hspace*{1em}+Monitor$_{\text{hidden}}$ & 98.9  & 100   \\
\bottomrule
\end{tabular}
}
\caption{
Hard Abstention (\%) of linear classifiers trained on HLE and tested on GSM8k and AIME.
}
\label{apptab:crossBenchmark}
\end{table}

In contrast, training on GSM8K is insufficient for developing Hard Abstention capability, because GSM8K contains very few unsolvable questions. The linear classifiers cannot reliably learn the hidden state patterns associated with unsolvability.

\section{Full-Dataset Trade-offs and Output Utility}
\label{app:fullTradeoff}

\paragraph{Full-dataset trade-offs.}
The main results report ACC over all questions and the remaining metrics over originally incorrect questions. For completeness, \autoref{apptab:fullTradeoff} reports ACC over all questions, HA $=P(\text{abstain}\mid\text{originally incorrect})$, FPA $=P(\text{abstain}\mid\text{originally correct})$, and average token usage over all questions. The monitors identify many originally incorrect cases while introducing few false-positive abstentions, and their full-dataset token reduction remains substantial.

\begin{table*}[h!]
\centering
\footnotesize
\setlength{\tabcolsep}{3.5pt}
\begin{tabular}{@{}llrrrrr@{}}
\toprule
\textbf{Model} & \textbf{Method} & \textbf{ACC $\uparrow$} & \textbf{HA $\uparrow$} & \textbf{FPA $\downarrow$} & \makecell{\textbf{Token (all) $\downarrow$}\\\textbf{2048}} & \makecell{\textbf{Token (all) $\downarrow$}\\\textbf{4096}} \\
\midrule
\multirow{4}{*}{GPT-oss-20B}
 & Original & 74.6 & 0.0 & 0.0 & 1813 & 3071 \\
 & +BoostAbstention & 74.6 (+0.0) & 0.0 (+0.0) & 0.0 (+0.0) & 1811 (-0.1\%) & 3050 (-0.7\%) \\
 & +Monitor$_{\text{express}}$ & \textbf{74.6 (+0.0)} & \textbf{100.0 (+100.0)} & \textbf{0.0 (+0.0)} & \textbf{1586 (-12.5\%)} & \textbf{2432 (-20.8\%)} \\
 & +Monitor$_{\text{hidden}}$ & 73.7 (-0.9) & 98.9 (+98.9) & 1.2 (+1.2) & 1750 (-3.5\%) & 2893 (-5.8\%) \\
\midrule
\multirow{4}{*}{R1-Distill-Qwen3-8B}
 & Original & 76.9 & 0.0 & 0.0 & 2040 & 4030 \\
 & +BoostAbstention & 76.9 (+0.0) & 0.0 (+0.0) & 0.0 (+0.0) & 2040 (-0.0\%) & 4030 (-0.0\%) \\
 & +Monitor$_{\text{express}}$ & 74.6 (-2.3) & 63.3 (+63.3) & 3.0 (+3.0) & 1903 (-6.7\%) & 3673 (-8.9\%) \\
 & +Monitor$_{\text{hidden}}$ & 73.8 (-3.1) & \textbf{97.9 (+97.9)} & 4.0 (+4.0) & \textbf{1748 (-14.3\%)} & \textbf{3371 (-16.4\%)} \\
\midrule
\multirow{4}{*}{R1-Distill-Qwen-32B}
 & Original & 56.2 & 0.0 & 0.0 & 1993 & 3649 \\
 & +BoostAbstention & 56.2 (+0.0) & 0.0 (+0.0) & 0.0 (+0.0) & 1974 (-1.0\%) & 3565 (-2.3\%) \\
 & +Monitor$_{\text{express}}$ & \textbf{56.2 (+0.0)} & \textbf{100.0 (+100.0)} & \textbf{0.0 (+0.0)} & 1447 (-27.4\%) & 2207 (-39.5\%) \\
 & +Monitor$_{\text{hidden}}$ & 55.3 (-0.9) & 97.0 (+97.0) & 1.6 (+1.6) & \textbf{1296 (-35.0\%)} & \textbf{2103 (-42.4\%)} \\
\midrule
\multirow{4}{*}{QwQ-32B}
 & Original & 71.5 & 0.0 & 0.0 & 2037 & 3917 \\
 & +BoostAbstention & 71.5 (+0.0) & 0.0 (+0.0) & 0.0 (+0.0) & 2037 (-0.0\%) & 3917 (-0.0\%) \\
 & +Monitor$_{\text{express}}$ & \textbf{71.5 (+0.0)} & \textbf{100.0 (+100.0)} & \textbf{0.0 (+0.0)} & 1785 (-12.4\%) & 3271 (-16.5\%) \\
 & +Monitor$_{\text{hidden}}$ & \textbf{71.5 (+0.0)} & 96.9 (+96.9) & \textbf{0.0 (+0.0)} & \textbf{1528 (-25.0\%)} & \textbf{2825 (-27.9\%)} \\
\bottomrule
\end{tabular}
\caption{Full-dataset accuracy--abstention--token trade-off (\%). Parentheses show changes from the original model; token columns report averages over all questions.}
\label{apptab:fullTradeoff}
\end{table*}

\paragraph{Output utility.}
We further evaluate whether the concise responses remain useful when a model is likely to fail. Following the example in \autoref{fig:optimizedAnswer}, we use GPT-5.5 as a rubric-based judge to compare the original long response with the concise response from the monitor on four five-point metrics: relevance, non-misleadingness, helpfulness, and readability. \autoref{apptab:outputUtility} shows that the concise responses provide more reliable and readable guidance, while preserving relevance.

\begin{table*}[h!]
\centering
\small
\setlength{\tabcolsep}{7pt}
\begin{tabular}{@{}llrrrr@{}}
\toprule
\textbf{Model} & \textbf{Output} & \textbf{Relevance $\uparrow$} & \textbf{Non-misleadingness $\uparrow$} & \textbf{Helpfulness $\uparrow$} & \textbf{Readability $\uparrow$} \\
\midrule
\multirow{2}{*}{GPT-oss-20B} & Original & 4.0 & 2.2 & 2.6 & 1.4 \\
 & +Monitor & 4.0 & 4.8 & 2.2 & 4.8 \\
\midrule
\multirow{2}{*}{R1-Distill-Qwen3-8B} & Original & 4.0 & 2.4 & 2.2 & 1.2 \\
 & +Monitor & 4.0 & 4.0 & 3.8 & 3.6 \\
\midrule
\multirow{2}{*}{R1-Distill-Qwen-32B} & Original & 4.2 & 2.6 & 2.4 & 1.6 \\
 & +Monitor & 4.4 & 4.2 & 4.2 & 4.6 \\
\midrule
\multirow{2}{*}{QwQ-32B} & Original & 4.0 & 2.4 & 2.0 & 1.2 \\
 & +Monitor & 4.2 & 3.6 & 4.0 & 4.6 \\
\bottomrule
\end{tabular}
\caption{Concise-output utility evaluation on a five-point scale.}
\label{apptab:outputUtility}
\end{table*}

\section{Analysis Beyond Mathematical Tasks}
\label{app:coding}

Since we capture unsolvability signals from reasoning expressions (\autoref{sec:cbReasoningTokens}) and hidden states (\autoref{sec:cbHiddenStates}) without relying on task-specific heuristics, they might generalize to other reasoning tasks. We evaluate this on LiveCodeBench and GPQA-Diamond.

\subsection{LiveCodeBench}

\begin{table*}[h!]
\centering
\small
\resizebox{0.75\linewidth}{!}{
\begin{tabular}{l|>{\centering\arraybackslash}p{2cm}>{\centering\arraybackslash}p{2cm}|>{\centering\arraybackslash}p{2cm}>{\centering\arraybackslash}p{2cm}}
\toprule
\multirow{2}{*}{\textbf{Model}}        & \multicolumn{2}{c|}{\textbf{Reasoning Expressions}} & \multicolumn{2}{c}{\textbf{Hidden States}} \\
& \textbf{ConfDiff} 
& \textbf{ConfCurv} 
& \textbf{LDA} 
& \textbf{LR} \\
\midrule
GPT-oss-20B          & 100   & 86.8 & 81.1 & 83.5 \\
R1-Distill-Qwen3-8B  & 97.3  & 93.0 & 92.0 & 93.9 \\
R1-Distill-Qwen-32B & 100   & 91.7 & 81.6 & 84.9 \\
QwQ-32B              & 100   & 92.5 & 77.3 & 85.4 \\
\bottomrule
\end{tabular}
}
\caption{
Accuracy (\%) of separating eventually solved and unsolved questions using signals related to the capability boundary, derived from reasoning expressions (at 2\% of the reasoning trace) and hidden states.}
\label{apptab:coding}
\end{table*}

We conduct experiments on 1055 samples from LiveCodeBench \citep{jain2024livecodebench}. \autoref{apptab:coding} shows the accuracy of separating solvable and unsolvable coding questions based on capability boundaries in reasoning expressions and hidden states.

We observe that \textit{coding tasks also exhibit clear capability boundaries in both reasoning expressions and hidden states}. However, the accuracy from hidden states is lower than that from mathematical tasks. A plausible explanation is that the difficulty of coding problems is less explicitly signaled in the input itself. Unlike in math exams where humans can often immediately recognize ``end-of-exam'' questions that they are unlikely to solve, coding tasks usually require some degree of initial reasoning before the model can detect unsolvability. As a result, the unsolvability signal in the hidden states of the last input token is weaker, whereas the signal becomes more apparent in the early reasoning expressions.

\subsection{GPQA-Diamond}

We additionally test whether the same signals separate originally correct from incorrect cases on GPQA-Diamond \citep{rein2023gpqa}. \autoref{apptab:gpqa} shows that the signal remains observable, although it is weaker than on the math tasks. GPQA is multiple-choice, so guessing can make final correctness a noisier proxy for genuine solvability than free-form math.

\begin{table}[H]
\centering
\small
\setlength{\tabcolsep}{5pt}
\begin{tabular}{@{}lrrrr@{}}
\toprule
\textbf{Model} & \textbf{ConfDiff} & \textbf{ConfCurv} & \textbf{LDA} & \textbf{LR} \\
\midrule
GPT-oss-20B & 71.2 & 60.0 & 81.2 & 87.5 \\
R1-Distill-Qwen3-8B & 87.5 & 91.2 & 93.8 & 87.5 \\
R1-Distill-Qwen-32B & 77.5 & 77.5 & 87.5 & 87.5 \\
QwQ-32B & 78.8 & 77.5 & 81.2 & 87.5 \\
\bottomrule
\end{tabular}
\caption{GPQA-Diamond correct/incorrect separation accuracy (\%).}
\label{apptab:gpqa}
\end{table}

\section{Monitoring Strategies with Maximum Context Length}
\label{app:maximumContextLength}

To evaluate whether our monitoring strategies (Monitor\(_{\text{express}}\) and Monitor\(_{\text{hidden}}\)) hold promise for optimizing long-context reasoning tasks \citep{ling2025longreason,kuratov2024babilong}, we provide the optimized token usage \textit{under extended token budget, set to the maximum} according to HuggingFace model card (128k for GPT-oss-20B, 64k for R1-Distill-Qwen3-8B, and 32k for both R1-Distill-Qwen-32B and QwQ-32B).

\begin{table}[H]
\centering
\small
\resizebox{0.99\linewidth}{!}{
\begin{tabular}{l|c|cc}
\toprule
                        & \textbf{Context Length} & \textbf{Token ↓} & \textbf{Overflow (\%) ↓}  \\

\midrule
GPT-oss-20B  & \multirow{3}{*}{128k}  &  $6023$ & $0$  \\
\hspace*{1em}+Monitor\(_{\text{express}}\) &    &  $1897$\diffdown{68.5}   &  $0$\\
\hspace*{1em}+Monitor\(_{\text{hidden}}\) &  & $3826$\diffdown{36.5} & $0$  \\
\midrule
R1-Distill-Qwen3-8B  & \multirow{3}{*}{64k}  &  $5096$ & $0$  \\
\hspace*{1em}+Monitor\(_{\text{express}}\) &    &  $2834$\diffdown{44.4}   &  $0$\\
\hspace*{1em}+Monitor\(_{\text{hidden}}\) &  & $2227$\diffdown{56.3} & $0$  \\
\midrule
R1-Distill-Qwen-32B & \multirow{3}{*}{32k}  &  $5910$ & $0$  \\
\hspace*{1em}+Monitor\(_{\text{express}}\) &    &  $1043$\diffdown{82.4}   &  $0$\\
\hspace*{1em}+Monitor\(_{\text{hidden}}\) &  & $1246$\diffdown{78.9} & $0$  \\
\midrule
QwQ-32B         &  \multirow{3}{*}{32k}  &  $5062$ & $0$  \\
\hspace*{1em}+Monitor\(_{\text{express}}\) &    &  $2002$\diffdown{60.5}   &  $0$\\
\hspace*{1em}+Monitor\(_{\text{hidden}}\) &  & $274$\diffdown{94.6} & $0$  \\
\bottomrule
\end{tabular}
}
\caption{
Monitoring strategies (Monitor\(_{\text{express}}\) and Monitor\(_{\text{hidden}}\)) with maximum context length.
}
\label{apptab:maximumContextLength}
\end{table}

Under these settings, \autoref{apptab:maximumContextLength} shows that the models always produce complete answers (Overflow = 0), and our monitoring strategies still \textit{reduce token usage significantly as expected} (Acc and HA are the same as under 2k or 4k context length).

Moreover, it is worth noting that \textit{our strategies are not constrained by context length}:
\begin{packeditemize}
\item Monitor\(_{\text{express}}\) continuously tracks the densities of confident and uncertain expressions throughout the reasoning process and can detect unsolvable questions \textit{at an early stage} (\autoref{fig:confDiffCurv}), well before the model reaches the end of reasoning.
\item Monitor\(_{\text{hidden}}\) identifies unsolvable questions directly from the hidden states of last input token, i.e., \textit{even before the model begins reasoning}.
\end{packeditemize}

Because our strategies operate early in the reasoning trajectory, we intentionally used relatively short context length (e.g., 2k or 4k) in \autoref{sec:reasoningWithCB} to demonstrate their early detection capabilities.

\section{Capability Boundary in Hidden States}
\label{app:CBhidden}

\begin{figure*}[t]
    \centering
    \includegraphics[width=0.99\linewidth]{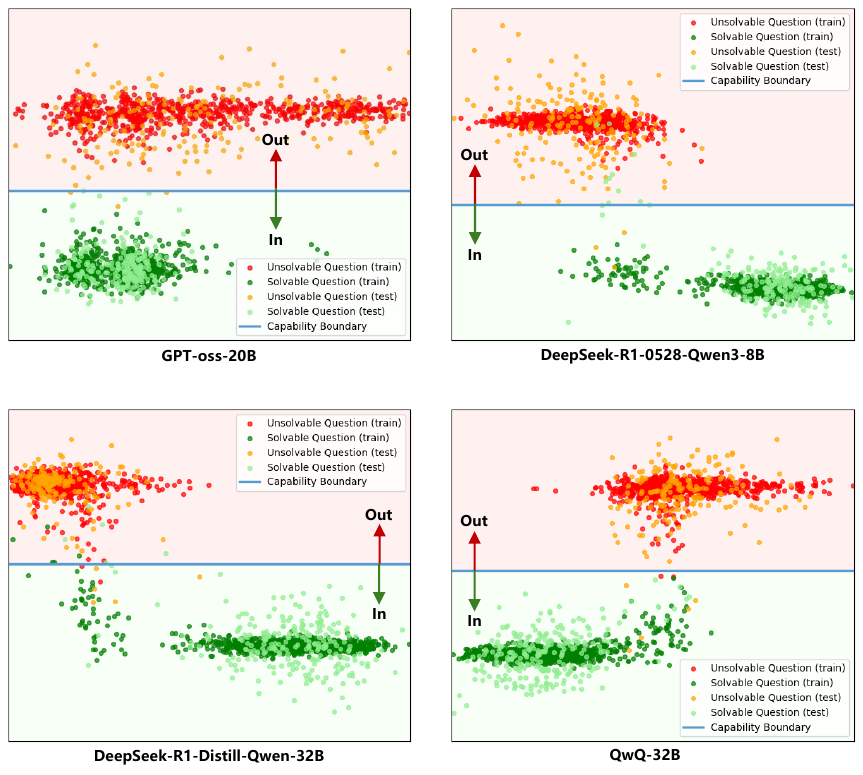}
    \caption{A linear classifier (e.g., LDA) clearly separates eventually solved and unsolved questions using hidden state signals related to the operational capability boundary.}
    \label{appfig:CBhidden}
\end{figure*}

\autoref{appfig:CBhidden} visualizes the classification of solvable and unsolvable questions of all LRMs, along with the capability boundary (blue line) determined by the decision boundary of linear classifier (e.g., LDA). We observe that the two classes of questions are clearly separated, under random train–test splits. These results indicate that the hidden states of the last input token contain useful signals related to the operational capability boundary, and is valid across different LRMs.

\section{Unproductive Reasoning}
\begin{figure*}
    \centering
    \includegraphics[width=0.9\linewidth]{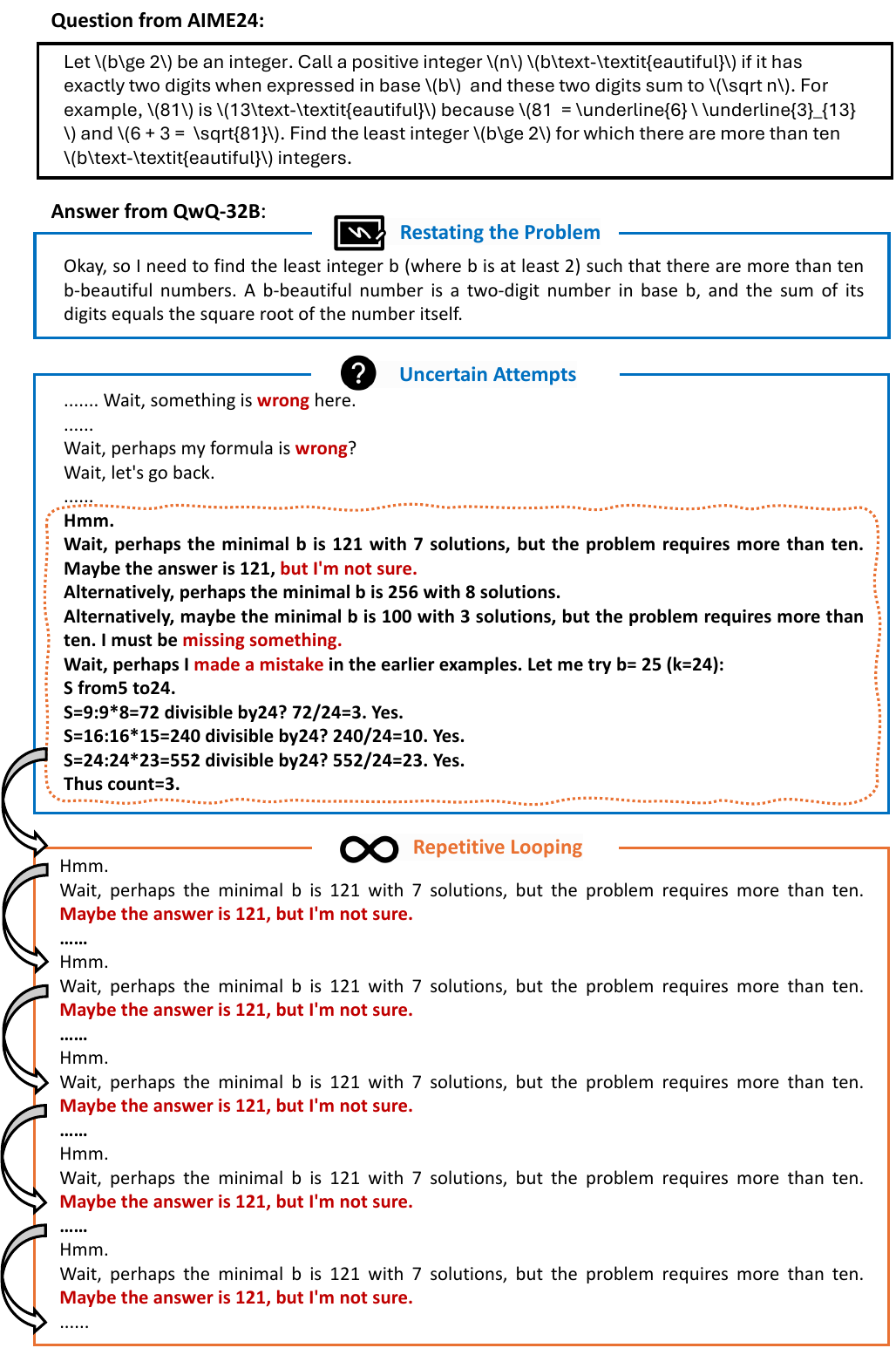}
    \caption{An example of unproductive reasoning manifested as repetitive looping in LRMs when attempting to solve a mathematical problem, where the model repeatedly cycles through the same uncertain deductions without making progress.}
    \label{fig:repetitive_looping}
\end{figure*}

\begin{figure*}
    \centering
    \includegraphics[width=0.9\linewidth]{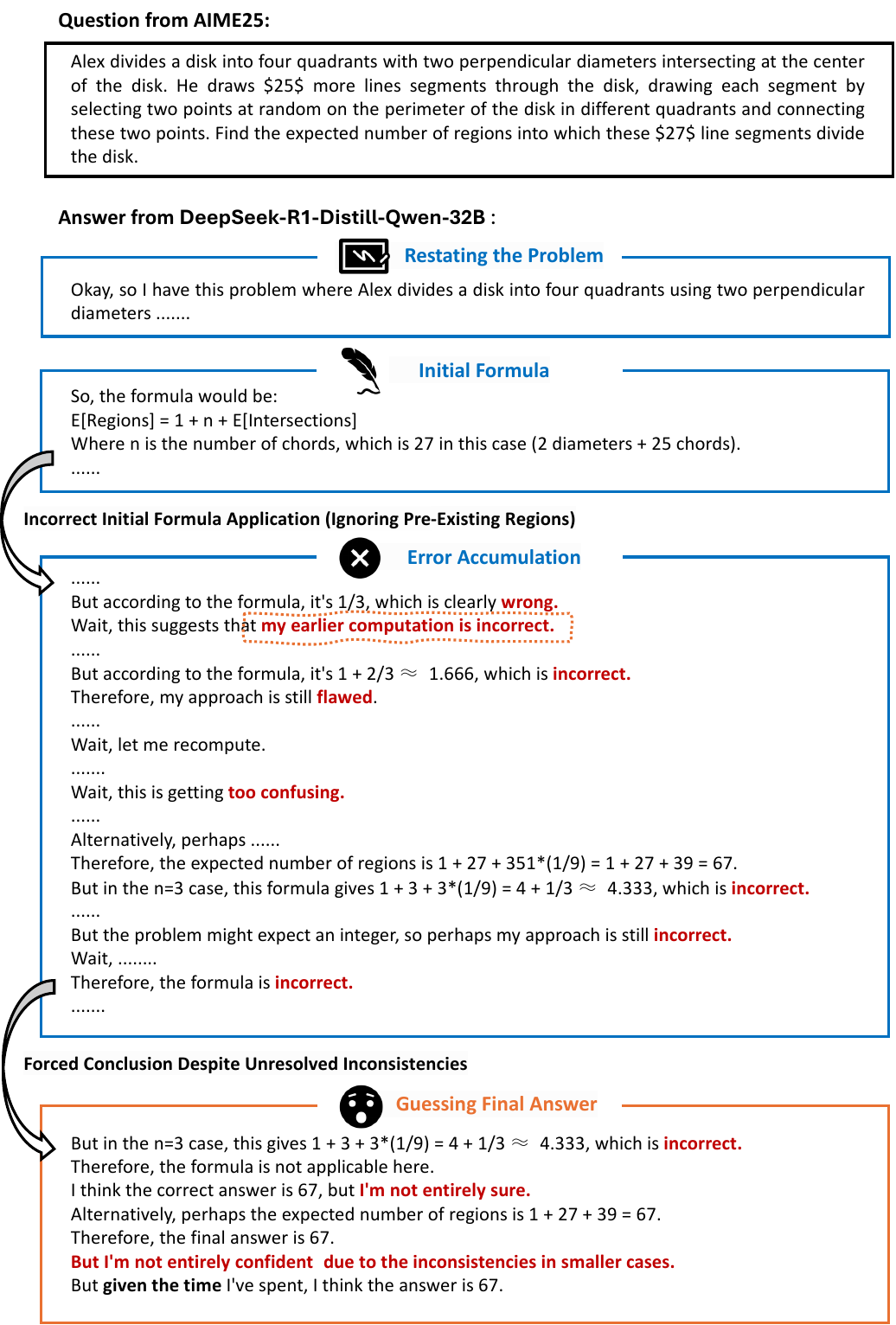}
    \caption{An example of error accumulation in LRMs when tackling a problem on the expected number of regions in a disk, where the model accumulates errors due to an incorrect initial formula, leading to inconsistent results and a forced answer guess despite unresolved contradictions.}
    \label{fig:error_accumulation}
\end{figure*}

\autoref{fig:repetitive_looping} and \autoref{fig:error_accumulation} show representative unproductive reasoning phenomena in LRMs, including repetitive looping and error accumulation, which impair their capacity for sound and consistent reasoning in complex problem-solving scenarios. They respectively show how models get stuck in repetitive deductions (\autoref{fig:repetitive_looping}) and how errors compound during formula application and result derivation (\autoref{fig:error_accumulation}). In mathematical problem reasoning, the former presents repeated deduction of the same uncertain conclusion, while the latter sees continuous error accumulation due to flawed initial methods.
\label{app:output}

\end{document}